\newif\ifarxiv

\arxivtrue

\ifarxiv
    \documentclass{article}
    \usepackage{fullpage}

\else
\documentclass[twoside]{article}
\usepackage[accepted]{aistats2021}
\fi

\usepackage{amsmath,amssymb,bm}
\usepackage{hyperref}
\usepackage{enumerate,mathtools}
\usepackage{amsthm}
\usepackage{microtype}

\usepackage{mathrsfs} 

\usepackage{caption}
\usepackage{subcaption}
\usepackage[round]{natbib}

\newtheorem{theorem}{Theorem}

\newtheorem{lemma}[theorem]{Lemma}

\theoremstyle{definition}
\newtheorem{definition}{Definition}
\newtheorem{remark}{Remark}

\newtheorem{example}{Example}

  \usepackage{pgfplots}
  \pgfplotsset{compat=newest}

\newcommand{\reals}{\mathbb{R}}
\newcommand{\bEx}{\ensuremath{\mathbb{E}}}

\newcommand{\ex}[1]{\ensuremath{\mathbb{E}\left[ #1\right]}}

\newcommand{\normal}{\mathcal{N}}

\DeclareMathOperator{\cov}{\mathsf Cov}

\newcommand{\cN}{\mathcal{N}}

\newcommand{\cP}{\mathcal{P}}
\newcommand{\cQ}{\mathcal{Q}}

\newcommand{\cT}{\mathcal{T}}

\newcommand{\eps}{\epsilon}

\usepackage{color}

\ifarxiv
\newcommand{\appname}{Appendix}
\newcommand{\appnamesmall}{appendix}
\else 
\newcommand{\appname}{Supplement}
\newcommand{\appnamesmall}{supplementary materials}
\fi

\usepackage{url}
\usepackage{ifthen}
\usepackage{bbm}

\let\originalleft\left
\let\originalright\right
\renewcommand{\left}{\mathopen{}\mathclose\bgroup\originalleft}
\renewcommand{\right}{\aftergroup\egroup\originalright}

\title{Convergence of  Gaussian-smoothed optimal transport distance with sub-gamma distributions and  dependent samples}

\usepackage{xcolor}

\ifarxiv

\author{Yixing Zhang\thanks{Department of Electrical and Computer Engineering, Duke University}  \and  Xiuyuan Cheng\thanks{Department of Mathematics, Duke University} \and Galen Reeves\thanks{Department of Electrical and Computer Engineering and the Department of Statistical Science, Duke University}}
\else
\runningtitle{Convergence of  GOT distance with sub-gamma distributions and  dependent samples}
\fi

\begin{document}

\ifarxiv
\maketitle

\else
\twocolumn[

\aistatstitle{Convergence of  Gaussian-smoothed optimal transport distance with sub-gamma distributions and  dependent samples}

\aistatsauthor{Yixing Zhang \And  Xiuyuan Cheng \And Galen Reeves  }

\aistatsaddress{Duke University \And  Duke University \And Duke University} ]

\fi

\begin{abstract}
The Gaussian-smoothed optimal transport (GOT) framework, recently proposed by Goldfeld et al.,  scales to high dimensions in estimation and provides an alternative to entropy regularization. This paper provides convergence guarantees for estimating the GOT distance under more general settings. For the Gaussian-smoothed $p$-Wasserstein distance in $d$ dimensions, our results require only the existence of a moment greater than $d + 2p$. For the special case of sub-gamma distributions, we quantify the dependence on the dimension $d$ and establish a phase transition with respect to the scale parameter. We also prove convergence for dependent samples, only requiring a condition on the pairwise dependence of the samples measured by the covariance of the feature map of a kernel space. 

A key step in our analysis is to show that the GOT distance is dominated by a family of kernel maximum mean discrepancy (MMD) distances with a kernel that depends on the cost function as well as the amount of Gaussian smoothing.  This insight provides further interpretability for the GOT framework and also introduces a class of kernel MMD distances with desirable properties.  The theoretical results are supported by numerical experiments.

\end{abstract}

\ifarxiv
\tableofcontents
\fi

\section{Introduction}

There has been significant interest in optimal transport (OT) distances for data analysis,
motivated by applications in statistics and machine learning
ranging from computer graphics and imaging processing \cite{solomon2014earth,ryu2018vector, li2018parallel}
to deep learning \cite{courty2016optimal,shen2017wasserstein, bhushan2018deepjdot}; see \cite{peyre:2019}. 
The OT cost between probability measures $P$ and $Q$  with cost function $c(x,y)$ is defined as
\begin{align}
\cT(P, Q) \coloneqq \inf _{\pi \in \Pi(P, Q)} \int c(x,y)\,  d \pi(x, y),
\end{align}
where $\Pi(P,Q)$ is the set of all probability measures whose marginals are $P$ and $Q$.
Of central importance to applications in statistics and machine learning is the rate at which the empirical measure $P_n$ of and iid sample approximates the true underlying distribution $P$.  
In this regard, one of the main challenges for OT distances is that rate convergence suffers from the curse of dimensionality: the number of samples $n$ needs to grow exponentially with the dimension $d$ of the data  \cite{fournier:2015}.

On a closely related note, OT also suffers from
 computational issues, particularly in the high-dimensional settings. To address both statistical and computation limitations, recent work has focused on regularized versions of OT including entropy regularization \cite{cuturi2013sinkhorn} and Gaussian-smoothed optimal transport (GOT) \cite{goldfeld:2020b}. 
The entropy-regularized OT has attracted intensive theoretical interest \cite{feydy2019interpolating,klatt2020empirical,bigot2019central},
as well as an abundance of algorithm developments \cite{gerber2017multiscale,abid2018greedy, chakrabarty2020better}.
In comparison, GOT is less understood both in theory and in computation. The goal of the current paper is thus to deepen the theoretical analysis of GOT under more general settings,
so as to lay a theoretical foundation for computational study and potential applications.

In particular, we consider distributions that satisfy only a bounded moment condition and general settings involving dependent samples. 
For the special case of sub-gamma distributions, we show a phase transition depending on the dimension $d$ and with respect to the scale parameter of the sub-gamma distribution. 
Going beyond the case of iid samples,
our convergence rate covers dependent samples as long as a condition on the pair-wise dependence quantified by the covariance of the kernel-space feature map is satisfied.
A key step in our analysis is to establish a novel connection between the GOT distance and a family of kernel MMD distances,
which can be of independent interest.
In the kernel MMD upper bound, 
the kernel is neither bounded nor translation invariant,
and is determined by both the cost function of OT and the amount of Gaussian smoothing. 
The theoretical findings are supported by numerical experiments.

To summarize our contribution, we provide an overview of the main theoretical results in the next subsection, and then close the introduction with a  detailed review of related work. After introducing notations and needed preliminaries in Section 2,
we derive upper bounds of GOT using kernel MMD of a new two-moment kernel in Section 3,
which leads to the convergence rate results in Section 4 and numerical results in Section 5. All proofs are in the \appnamesmall.

\subsection{Overview of Main Results} 
 
In this paper, we focus on the OT cost $\cT_p(P,Q)$ associated with the cost function $c_p(x,y) = \|x- y\|^p$ for $p > 0$ and $c_0(x,y)= 1_{\{x \ne y\}}$. The total variation distance is given by  $\cT_0(P,Q)$ and the $p$-Wasserstein distance is given by $\cT_p(P,Q)$ if $p \in (0,1]$ and $(\cT_p(P,Q))^{1/p}$ if $p > 1$~\cite{villani2003topics}.

The minimax convergence rate of $\cT_p(P,P_n)$ was established by 
\cite[Theorem~1]{fournier:2015} who showed that if $P$ has a moment strictly greater than $2p$, then  
\begin{align}  
    \ex{ \cT_{p} (P, P_n) } \asymp  \begin{cases} n^{-\frac{1}{2}}, & p > d/2\\
    n^{-\frac{1}{2}} \log n , & p = d/2\\
    n^{- \frac{p}{d}},  & p \in (0,  d/2)
    \end{cases}. \label{eq_Tp_rate}
\end{align}
Unfortunately, this means that the sample complexity increases exponentially with the dimension for $d > 2p$.

Recently, \cite{goldfeld:2020b} showed that one way to overcome the curse of dimensionality is to consider the Gaussian-smoothed OT distance, defined as
\begin{align*}
    \cT_p^{(\sigma)}(P,Q) \coloneqq  \cT_p( P \ast \normal_\sigma, Q \ast \normal_\sigma),
\end{align*}
 where $\cN_\sigma$ denotes the iid Gaussian measure with mean zero and variance $\sigma^2$. Under the assumption that $P$ is sub-Gaussian with constant $v$, they proved an upper bound on the converge rate that is independent of the dimension:
 \begin{align*}
\ex{ \cT_{p} (P, P_n) } \le \frac{C_{d,p,\sigma,v}}{\sqrt{n}}, \quad p \in \{0,1,2\}.
\end{align*}
The precise the form of the constant $C_{d,p,\sigma,v}$ is provided for $p \in \{0,1\}$ but not for the case $p =2$ unless $P$ is also assumed to have bounded support.
Ensuing work by \cite{goldfeld2020asymptotic} established the same convergence rate for $p=1$ under the relaxed assumption that $P$ has finite moment grater than $2d + 2$. 

Metric properties of GOT were studied by \cite{goldfeld:2020} who showed that   $\cT_1^{(\sigma)}(P,Q)$ is a metric on the space of probability measures with finite first moment and that the sequence of optimal couplings converges in the $\sigma \to 0$ limit to the optimal coupling for the unsmoothed Wasserstein distance. Their arguments depend only on the pointwise convergence of the characteristic functions under Gaussian smoothing, and thus also apply to the case of general $p$ considered in this paper. 

One of the main contributions of this paper is to prove an upper bound on the convergence rate for all orders of $p$ and under more general assumptions on $P$. Specifically, we prove the following result:

\begin{theorem}\label{thm_GOT_moment_UB} Let $P_n$ be the empirical measure of $n$ iid samples from a probability measure $P$ on $\reals^d$ that satisfies $(\int \|x\|^s \, dP(x))^{1/s} \le m$ for some $s > d + 2p$. There exists a positive constant $C_{d,p,s}$ such that for all $\sigma > 0$, 
 \begin{align}
\ex{ \cT^{(\sigma)}_{p} (P, P_n) } \le C_{d,p,s} \frac{  \sigma^p}{\sqrt{n}}  \left( 1 + \frac{m}{\sigma} \right)^{\frac{d}{2} +p}.
\end{align}
\end{theorem}

This result brings the GOT framework in line with the general setting studied by \cite[Theorem~1]{fournier:2015}, and shows that the benefits obtained by smoothing extend  beyond the special cases of small $p$ and well-controlled tails. To help interpret this result it is important to keep in mind that for $p> 1$, the Wasserstein distance is given by the $p$-th root of the GOT. As for the tightness of the bound, there are two regimes worth considering,
namely when $\sigma\to 0$  as $n\to \infty$ and when $\sigma$ is fixed.
In the former case, the dependence on $\sigma$ seems to be nearly tight. In Section~\ref{sec_kxx}, we show that if $\sigma \asymp n^{-\frac{1}{ d+ 2p}}$ then Theorem~\ref{thm_GOT_moment_UB} implies an upper bound on the \emph{unsmoothed} convergence rate 
\begin{align}
\ex{ \cT_{p} (P, P_n) } \le C_{d,p,s} m^p n^{-\frac{p}{d + 2p}}. \label{eq:OT_near_optimal}
\end{align}
Notice that for $d \ll 2p$ and $d \gg 2p$ this recovers the minimax convergence rate given in \eqref{eq_Tp_rate}.

The main technical step in our approach is to establish a novel connection between GOT and a family of kernel MMD distances (Theorem~\ref{thm_general_UB}). We then show how a particular member of this family, which we call the `two-moment' kernel,  defines a metric on the space of probability measures with finite moments strictly greater than $p + d/2$ (Theorem~\ref{thm_two_moment}). 

In addition to Theorem~\ref{thm_GOT_moment_UB}, we also provide  further results that elucidate the role of the dimension as well the tail behavior of the underlying distribution (Theorem~\ref{thm_kbound_subgamma}). Furthermore, we address the setting of dependent samples and provide an example illustrating how the connection with MMD can be used to go beyond the usual assumptions involving mixing conditions for stationary processes. Finally, we provide some numerical experiments that support our theory.

\subsection{Comparison with Previous Work}
The convergence of OT distances continues to be an active area of research \cite{singh:2018,jonathan2019minmax,lei2020convergence}. Building upon the the work of work of \cite{cuturi2013sinkhorn}, a recent line of work has focused on entropy regularized OT defined by 
\begin{align*}
S_\eps(P,Q) \coloneqq \inf_{\pi \in \Pi(P,Q)} \int c(x,y) \, d\pi(x,y) + \epsilon D(\pi \| P \otimes Q)
\end{align*}
where $D(\mu \| \nu)= \int  \log \frac{ d \mu}{ d \nu} d \mu$ is the relative entropy between probability measures $\mu$ and $\nu$. The addition of the regularization term facilitates the numerical approximation using the  Sinkhorn algorithm. The amount of regularization interpolates between OT in the $\eps\to 0$ limit and the kernel MMD in $\eps\to \infty$ limit; see \cite{feydy2019interpolating}. In contrast to the Gaussian-smoothed Wasserstein distance, entropy regularized OT is not a metric since it does not satisfy the triangle inequality. Convergence rates for entropy regularized OT were obtained by  \cite{genevay2019sample} under the assumption of bounded support and more recently by \cite{mena:2019} under the assumption of sub-Gaussian tails.  Further properties have been studied by \cite{luise2018differential} and \cite{klatt2020empirical}.

There has also been work focusing on the sliced Wasserstein distance,  which is obtained by averaging the one-dimensional Wasserstein distance over the unit sphere \cite{rabin2011wasserstein,bonneel2015sliced}. While the sliced Wasserstein distance is equivalent to the Wasserstein distance in the sense that convergence in one metric implies convergence in another, the rates of convergence need not be the same. See Section 1.2 of \cite{goldfeld2020asymptotic} for further discussion.

Going beyond convergence rates for empirical measures, properties of smoothed empirical measures have been studied in a variety of contexts, including the high noise limit \cite{chen2020asymptotics} and applications to the estimation of mutual information in deep networks \cite{goldfeld2018estimating}. Finally, we note that there has also been some work on convergence with dependent samples by \cite{fournier:2015}, who focus on OT distance, and also by \cite{young2019consistent}, who  consider a closely related entropy estimation problem.

\section{Preliminaries}

Let $\cP(\reals^d)$ be the space of Borel probability measures on $\reals^d$ and let $\cP_s(\reals^d)$ be the space of probability measures with finite $s$-th moment, i.e, $\int \|x\|^s \,dP(x) < \infty$. The Gaussian measure on $\reals^d$ with mean-zero and covariance $\sigma^2 I_d$ is denoted by $\cN_\sigma$. The convolution of probability measures $P$ and $Q$ is denoted by $P \ast Q$.  The Gamma function is given by $\Gamma (z)=\int _{0}^{\infty }x^{z-1}e^{-x}\,dx$ for $z > 0$. We use $C$ to denote a generic positive real number, and the value of $C$ may change from place to place.

\paragraph{Kernel MMD.}
A symmetric function $k\colon \reals^d \times \reals^d \rightarrow \reals$ is said to be a \emph{positive-definite kernel} on $\reals^d$ if and only if  for every $x_1, \dots, x_n \in \reals^d$, the symmetrix matrix $(k(x_i, x_j))_{i,j=1}^n$ is positive semidefinite.  A positive definite kernel $k$ can be used to define a distance on probability measure
known as RKHS MMD \cite{anderson1994two, gretton2005measuring, smola2007hilbert, gretton2012kernel}.
 Let $\cP^k(\reals^d)$ be the space of probability measures such that $\int \sqrt{k(x,x)} d P(x) < \infty $. For $P, Q \in \cP^k(\reals^d)$ the kernel MMD distance is defined as
\begin{align*}
    \gamma^2_k(P,Q) = \iint k(x,y) \, d(P(x) - Q(x))\, d(P(y) - Q(y)) .
\end{align*}
The  distance $\gamma_k(P,Q)$ is a pseudo-metric in general.  A kernel $k$ is said to be \emph{characteristic} to a set $Q \subseteq\cP$ of probability measures if and only if $\gamma_k$ is a metric on $\cQ$ \cite{gretton2012kernel,sriperumbudur2010hilbert}.
An alternative representation is given by
\begin{align}
    \gamma^2_k(P,Q) = \ex{ k(X,X')} + \ex{k(Y,Y')} - 2 \ex{ k(X,Y)} \label{eq_gamma_k_alt}
\end{align}
where $X,X'$  $\sim P$  iid, and $Y,Y'$ $\sim Q$ iid. The kernel MMD distance was shown  to be equivalent to energy distance in \cite{sejdinovic2013equivalence},
and a variant form 
used in practice is the kernel mean embedding statistics \cite{8187176, chwialkowski2015fast, jitkrittum2016interpretable}. 
Another appealing theoretical property of the kernel MMD distance is
its representation via the spectral decomposition of the kernel \cite{epps1986omnibus, fernandez2008test},
which gives rise to estimation consistency as well as practical algorithms of computing \cite{zhao2015fastmmd}.  
Kernel MMD has been widely applied in  data analysis and machine learning, including independence testing  \cite{fukumizu2008kernel,zhang2012kernel}
and generative modeling \cite{li2015generative,li2017mmd}.

\paragraph{Magnitude of multivariate Gaussian.} 
Our results also depend on some properties of the noncentral chi distribution. Let $Z \sim \normal(\mu,I_d)$ be a Gaussian vector with mean $\mu$ and identity covariance. The random variable $X = \|Z\|$ has chi-distribution with parameter $u= \|\mu\|$. The density is given by
\begin{align}
    g_{d,u}(x) =\frac{e^{-(x^2+u^2)/2} x^{d-1}}{2^{\frac{d}{2} -1}}  \sum _{k=0}^{\infty }\frac{(ux/2)^{2k}}{k! \Gamma \left({\frac {d+2k}{2}}\right)}.
    \label{eq_chi_density}
\end{align}
The $s$-th moment of this distribution is denoted by  $M_{d,u}(s) = \int x^s \, g_{d,u}(x) \, dx$. This function is an even polynomial of degree $s$ whenever $s$ is an even integer (see \appname~\ref{sec_kernel_numerical_computation}).  The special case $u=0$ corresponds to the (central) chi distribution and is given by $M_{d} (s) \coloneqq  2^\frac{s}{2} \Gamma(\tfrac{d+s}{2}) / \Gamma(\tfrac{d}{2}).$

\section{Upper Bounds on GOT} 

In this section, we show that GOT is bounded from above by a family of kernel MMD distances. It is assumed throughout that $d$ is a positive integer, $p \in (0, \infty)$, and $\sigma \in (0, \infty)$.

\subsection{General Bound via Kernel MMD}
Consider the feature map $\psi_x \colon \reals^d \to [0, \infty)$ defined by
\begin{align}
    \psi_x(z) &\coloneqq \frac{ \sqrt{\omega_d}}{ 2^\frac{d+p}{2}} \frac{ \|z\|^\frac{d-1  + 2p}{2}}{ \sqrt{ f(\|z\|)}} \phi\left(\frac{ z}{\sqrt{2}}  - \frac{ x}{\sigma} \right ),
\end{align}
where  $\omega_d = 2\pi^{d/2} / \Gamma(d/2)$ is the volume of the unit sphere in $\reals^d$, $\phi(u) = (2\pi)^{-d/2} \exp( -\frac{1}{2} \|u\|^2)$ is the standard Gaussian density on $\reals^d$, and $f$ is a probability density function on $[0, \infty)$ that satisfies 
\begin{align}
f(x) \ge a x^{d + 2p -1}  \exp( - b  x^2),
\label{eq:fLB}
\end{align}
for some $a > 0$ and $b \in (0,1/2)$. 
This feature map defines a positive semidefinite kernel $k$ according to
\begin{align*}
    k(x,y) = \langle \psi_x, \psi_y \rangle.
\end{align*}
After some straightforward manipulations (see \appname~\ref{sec_proof_UB}), one finds that $k(x,y)$ is finite on $\reals^d \times \reals^d$ and can be expressed as the product of a Gaussian kernel and a term that depends only on $\|x+y\|$. Specifically, 
\begin{align}
     k(x,y) & = \exp\left( -\frac{ \|x-y\|^2}{ 4 \sigma^2}\right) I_f\left( \frac{ \|x+y\|}{\sqrt{2} \sigma} \right), \label{eq_kernel_f}
\end{align}
where
\begin{align}
    I_f(u) &\coloneqq \frac{ \omega_d}{2^{d + p} ( 2\pi)^{\frac{d}{2}} }\int_0^\infty  \frac{ x^{d-1+2p}}{ f(x)} g_{d,u}\left(x \right)   \, dx, \label{eq:If} 
\end{align}
and $g_{d,u}(x)$ is the density of the non-central chi-distribution given in \eqref{eq_chi_density}. Note that this kernel is not shift invariant because of the term $I_f(u)$. 

The next result shows that the MMD defined by this kernel provides an upper bound on the GOT.  

\begin{theorem}\label{thm_general_UB}
Let $k$ be defined as in \eqref{eq_kernel_f}. For any $P, Q \in \cP(\reals^d)$ such that $\int \sqrt{k(x,x)} \, dP(x)$ and  $\int \sqrt{k(x,x)} \, dQ(x)$ are finite, the MMD defined by $k$ provides and upper bound on the GOT:
\begin{align*}
\cT^{(\sigma)}_p(P,Q) \le  2^{\max(p-1,0)} \sigma^p \gamma_k(P,Q).
\end{align*}
\end{theorem}

The significance of Theorem~\ref{thm_general_UB} is twofold. From the perspective of GOT, it provides a natural connection between the role of Gaussian smoothing and normalization of the kernel. From the perspective of MMD, Theorem~\ref{thm_general_UB} describes a family of kernels that metrize convergence in distribution as well as convergence in $p$-th moments.

Similar to the analysis of convergence rates in previous work \cite{goldfeld:2020,  goldfeld:2020b},  the proof of Theorem~\ref{thm_general_UB} builds upon  the fact that $\cT_p(P,Q)$ can be upper bounded by a weighted total variation distance~\cite[Theorem~6.13]{villani2008optimal}. The novelty of Theorem~\ref{thm_general_UB} is that it establishes an explicit relationship with the kernel MMD and also provides a much broader class of upper bounds parameterized by the density $f$.

\subsection{A `Two-moment' Kernel}

One potential limitation of Theorem~\ref{thm_general_UB} is that for a particular choice of density $f$, the requirement that $\sqrt{k(x,x)}$ is integrable might not be satisfied for probability measures of interest. For example, the convergence rates in  \cite{goldfeld:2020} and \cite{goldfeld:2020b} can be obtained as a corollary of  Theorem~\ref{thm_general_UB} by choosing $f$ to be the density of the generalized gamma distribution.
However, the inverse of this density grows faster than exponentially, and as a consequence, the resulting bound can be applied only to the case of sub-Gaussian distributions. 

The main idea underlying the approach in this section is that choosing a density with heavier tails leads to an upper bound that holds for a larger class of probability measures. Motivated by the functional inequalities appearing in \cite{reeves:2020d}, we consider the following density function, which belongs to the family of generalized beta-prime distributions:
\begin{align*}
f(x)= \frac{\eps}{2\pi x} \left( \left(\frac{x}{\lambda}\right)^{-\eps} + \left(\frac{x}{\lambda}\right)^{\eps}  \right)^{-1}. 
\end{align*}
For this special choice, the function $I_f(u)$ can be expressed as the weighted sum of two moments of the non-central chi distribution. Starting with  \eqref{eq_kernel_f} and simplifying terms leads to the following:

\begin{definition}\label{def:two_moment} The \emph{two-moment} kernel $k \colon \reals^d \times \reals^d \to \reals$ is defined as
 \begin{align}
    k(x,y)  &= \alpha_{d,p} \exp\left( -\frac{ \|x-y\|^2}{ 4 \sigma^2}\right) J\left( \frac{ \|x+y\|}{\sqrt{2} \sigma} \right) \label{eq:kernel_two_moment}
\end{align}
for all $\eps \in (0,d+2p]$ and $\lambda \in (0,\infty)$, where 
\begin{align*}
\alpha_{d,p} &\coloneqq  \frac{(2 \pi) 2^{-(p+d)} 2^{-d/2} }{\Gamma(\frac{d}{2})}\\
    J(u)& \coloneqq  \frac{  \lambda^\eps M_{d,u}(d \!+\! 2p \!-\! \eps) +  \lambda^{-\eps} M_{d,u}(d\!+\!2p \!+\! \eps) }{ 2 \eps}.
\end{align*}
In this expression,  $M_{d,u}(s)$ denotes the $s$-th moment of the non-central chi distribution with  $d$ degrees of freedom and parameter $u$.
\end{definition}

A useful property of the two-moment kernel is that it satisfies the upper bound
\begin{align*}
k(x,y) \le C_{d,p,\eps,\lambda} \left(  1 + \|x\|^{d + 2p + \eps} + \|y\|^{d + 2p + \eps}\right),
\end{align*}
where the constant depends only on $(d,p,\eps, \lambda)$ (see \appname~\ref{sec_proof_UB} for details). As a consequence:

\begin{theorem}\label{thm_two_moment}
Fix any $s > p + d/2$. For all $0 < \eps < \min(d + 2p, 2s  - 2p - d) $ and $\lambda \in (0, \infty)$ the  MMD defined by the two-moment kernel is a metric on the space $\cP_s(\reals^d)$ of probability measures with finite $s$-th moment. Furthermore, for all $P, Q \in \cP_s(\reals^d)$, 
\begin{align*}
\cT_p^{(\sigma)}(P,Q)\le  2^{\max(p-1,0)} \sigma^p  \gamma_k(P, Q).
\end{align*}
\end{theorem}

\begin{remark}
If $d + 2p \pm \eps$ are even integers then $J(u)$ is an even polynomial of degree $s= d + 2p + \eps$ with non-negative coefficients. For example,  if $d = 3$, $p =1$, and  $(\eps, \lambda) = (1,1)$, then
\begin{align}
J(u)
& = 60 + \frac{ 115}{2} u^2 + 11 u^4 + \frac{1}{2} u^6.
\end{align}
Methods for efficient numerical approximation of $J(u)$ are provided in \appname~\ref{sec_kernel_numerical_computation}.
\end{remark}

\section{Convergence Rate}\label{sec:convergence}

We now turn our attention to the fundamental question of how well the empirical measure of iid samples approximates the true underlying distribution. Let $S_1, \dots, S_n \in\reals^n$ be a sequence of $n$ independent samples with common distribution $P$. The empirical measure $P_n$ is the (random) probability measure on $\reals^d$ that places probability  mass $1/n$ at each sample point:
\begin{equation}\label{eq:def-Pn}
P_n := \frac{1}{n} \sum_{i=1}^n \delta_{S_i},
\end{equation}
where $\delta_x$ denotes the pointmass distribution at $x$. 

Distributional properties of the kernel MMD between $P$ and $P_n$ have been studied extensively \cite{gretton2012kernel}. For the purposes of this paper, we will focus on the expected difference between these distributions. As a straightforward consequence of  \eqref{eq_gamma_k_alt} one obtains an exact expression for the expectation of the squared MMD: 
\begin{align}
  \bEx\left[ \gamma^2_k(P, P_n)  \right] & =\frac{\bEx[k(X, X)]  - \bEx[k(X, X')]}{n}, \label{eq:gamma_k_moment}
\end{align}
where $X,X'$ are independent draws from $P$. Note that the numerator can also be expressed as $\ex{ \gamma_1^2(P, P_1)}$, i.e., the squared MMD under $n=1$ samples. By Jensen's inequality, the first moment satisfies 
\begin{align}
  \bEx\left[ \gamma_k(P, P_n)  \right] &\le \frac{\sqrt{\bEx[k(X, X)]}}{\sqrt{n}}. \label{eq:gamma_k_first_moment}
\end{align}
In the following, we focus on the two-moment kernel given in \eqref{eq:kernel_two_moment} and study how $\bEx[k(X, X)]$ depends on $P$ and the parameters $(p,\sigma)$.

We note that because $\gamma_k$ satisfies the triangle inequality, all of the results provided here extend naturally to the two-sample settings where one the goal is to approximate the distance $\gamma_k(P,Q)$ based on the empirical measures $P_n$ and $Q_m$.

\subsection{Finite Moment Condition} \label{sec_kxx}

We begin with an upper bound on $\bEx[k(X, X)]$ that holds whenever $\|X\|$ has a moment greater than $d + 2p$. 

\begin{theorem}\label{thm:HOmoment}
Let $X \in \reals^d$  be a random vector that satisfies $(\ex{ \|X\|^s})^{1/s}  \le m$  for some $s > d + 2p$. If $k$ is the two-moment kernel given in \eqref{eq:kernel_two_moment} with parameters $0 < \eps \le  \min(d +2p, s - d - 2p)$ and  $\lambda =  \sqrt{d+2p + \eps}    +   \sqrt{2} m/ \sigma $,  then
\begin{align*}
 \bEx[ k(X,X)] &\le \frac{\alpha_{d,p}}{\eps} \Big(\sqrt{2d+2p + \eps}    +   \frac{\sqrt{2} m}{\sigma} \Big)^{d + 2p}  .
 \end{align*}
\end{theorem}

In view of \eqref{eq:gamma_k_first_moment}, Theorem~\ref{thm_GOT_moment_UB} follows as an immediate corollary.

\paragraph{Small noise limit and unsmoothed OT.} It is instructive to consider the implications of Theorem~\ref{thm_GOT_moment_UB} in the limit  as $\sigma$ converges to zero. By two applications of the triangle inequality and the fact that $\cT_p(Q, Q\ast\normal_\sigma) \le \sigma M_{d}(p)$ for every $Q \in \cP$,  one finds that, for any $\sigma \in [0, \infty)$,  the (unsmoothed) OT distance can be upper bounded according to
\begin{align}
    \cT_p(P,P_n) \le C_{d,p} \left( \cT^{(\sigma)}_p(P,P_n)  + \sigma^p \right). \label{eq:Tp_triangle}
\end{align} 
Combining  \eqref{eq:Tp_triangle} with Theorem~\ref{thm_GOT_moment_UB} and then evaluating  at $\sigma =  m  n^{-\frac{1}{d + 2p}}$ leads to the (unsmoothed) convergence rate given in \eqref{eq:OT_near_optimal}.

\subsection{Sub-gamma Condition} 
Next, we provide a refined bound for distributions satisfying a sub-gamma tail condition. 
\begin{definition}
 A random vector $X \in \reals^d$ is said to be \emph{sub-gamma} with variance parameter $v > 0$ and scale parameter $b \ge 0$ if 
\begin{align}
\ex{ \exp( \alpha^\top X)} \le \exp\left( \frac{ v \| \alpha\|^2 }{2 ( 1- b \|\alpha\|)} \right) 
\end{align}
for all $\alpha \in \reals^d$ such that $\|\alpha\| \le 1/b$. If this condition holds with $b =0$ then $X$ is said to be \emph{sub-Gaussian} with variance parameter $v$. 
\end{definition}

Properties of sub-Gaussian and sub-gamma distributions have been studied extensively; see e.g, ~\cite{boucheron2013concentration}. In particular,  if  $X_1$ and $X_2$ are independent sub-gamma random vectors with parameters $(v_1, b_1)$ and $(v_2, b_2)$, respectively, then $X_1 +X_2$ is  sub-gamma with parameters $(v_1 + v_2, \max(b_1, b_2))$. 

The next result provides an upper bound on the moments of the magnitude of a sub-gamma vector. Although there is a rich literature this topic, we were unable to find a previous statement of this result and so it may be of independent interest. 

\begin{theorem}\label{lem:subgamma_bnd}
Let $X \in \reals^d$ be a sub-gamma random vector with parameters $(v, b)$. For all $s \in [0, \infty)$ 
\begin{align}
\ex{ \|X\|^s}  & \le  \sqrt{ 2 e} \left(  \sqrt{v}  + \sqrt{s}\, b  \right)^s  M_d(s),
\end{align} 
where  $M_{d} (s) :=  2^\frac{s}{2} \Gamma(\frac{d+s}{2}) / \Gamma(\frac{d}{2})$ is the $s$-th moment of the chi distribution with $d$ degrees of freedom. Furthermore, $M_d(s) \le \overline{M}_d(s)$ where
\begin{align}
\overline{M}_d(s)
&: = \left( \frac{ d + s}{e} \right)^\frac{s}{2} \left( \frac{ d + s}{ d} \right)^\frac{d-1}{2}. \label{eq:Mbar}
\end{align} 
\end{theorem}

\begin{theorem}\label{thm_kbound_subgamma} 
Let $X \in \reals^d$ be a sub-gamma random vector with parameters $(v, b)$. Let $r = d + 2p$ and for $t \in (0, d + 2p]$ define
\begin{align*} 
m(t) &\coloneqq (  \sqrt{\sigma^2 + 2 v} +  \sqrt{r + t} \,b  )^{r + t}  \overline{M}_{d}(r +  t),
\end{align*}
where $\overline{M}_d(s)$ is defined in \eqref{eq:Mbar}. If  $k$ is the two-moment kernel defined in \eqref{eq:kernel_two_moment} with  parameters $\eps = \sqrt{d}$ and $\lambda = (m(\eps)  / m(-\eps) )^{1/(2 \eps)}$  then
\begin{align}
 \ex{ k(X,X)}
 & \le  C (d + p)^p \! \Big(  \sqrt{1 +  \frac{2v}{\sigma^2}} +   \frac{\sqrt{d+2p} \,b}{\sigma}  \Big)^{d+2p}    ,
 \label{eq_ub-cor-subgamma}
 \end{align}
 where $C= \sqrt{2 \pi}  e^{7/8} < 6.02$.
\end{theorem}

In view of  \eqref{eq:gamma_k_first_moment},  Theorem~\ref{thm_kbound_subgamma} gives an upper bound on the convergence rate of $\gamma_k(P,P_n)$ with an explicit dependence on the sub-gamma parameters $(d,p,\sigma,v,b)$. 
For the special case of a sub-Gaussian distribution ($b =0$) and $p \in \{0,1\}$, this bound recovers the results 
in \cite{goldfeld:2020}. Going beyond the setting of sub-Gaussian distributions (i.e., $b > 0$) this bound quantifies the dependence on the dimension $d$ and the scale parameter $b$.

\paragraph{Phase transition in scale parameter.}
In the high-dimensional setting where $d$ increases with $n$, Theorem~\ref{thm_kbound_subgamma}  exhibits two distinct regimes depending on the behavior of the scale parameter. Suppose that $(p, \sigma, v)$ are fixed while $b$ scales with $d$.  If $b = O( d^{- \delta})$ for some  $\delta  > 1/2$  then $\bEx[k(X,X)]$ grows at most exponentially with the dimension:
\begin{align}
 \frac{\log  \bEx[k(X,X)]}{d}  \le  \frac{1}{2}\log\left( 1 +  \frac{2v}{\sigma^2} \right) + o(1).
\end{align}
Conversely if $b = \Omega(d^{-\delta})$  for $\delta <1/2$ then the upper bound increases faster than exponentially:
\begin{align*}
  \frac{ \log \bEx[k(X,X)]}{ d  } \le  \frac{2 (1 - \delta)  \log d}{2}  + O(1) .
\end{align*}

The following example provides a specific example of a sub-gamma distribution which shows that the upper bound in  Theorem~\ref{thm_kbound_subgamma} is tight with respect to the scaling of the dimension $d$ and the scale parameter $b$. Full details of this example are provided in \appname~\ref{sec:example_sub_gamma}.

\begin{example}\label{eg:1}
Suppose that $X = \sqrt{U}Z$ where $Z \sim \normal(0, I_d)$ is a standard Gaussian vector and $U$ is an independent Gamma random variable with shape parameter $1/(2 b^2)$ and scale parameter $2 b^2$. Then, $X$ satisfies the sub-gamma condition with parameters $(1,b)$ and so the upper bound in Theorem~\ref{thm_kbound_subgamma} applies. Moreover, for every pair $(\eps,\lambda)$ the expectation of the two-moment kernel satisfies the following lower bound 
\begin{align}
\ex{k(X,X)} &\ge \frac{ \alpha_{d,p}  }{\eps } \left(\frac{\sqrt{2} b}{\sigma} \right)^{r} \left( M_{b^{-2}}(r - \eps)M_{b^{-2}}(r + \eps)\right)^{1/2} \notag\\
& \quad \times\left( M_{d}(r - \eps)M_{d}(r + \eps)\right)^{1/2}. \label{eq_kxx_example_LB}
\end{align}
\end{example}

\begin{figure*}
\centering
  \includegraphics[width=0.65 \textwidth]{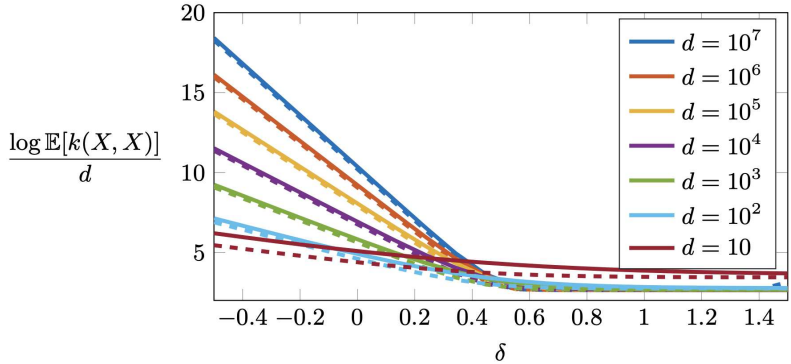}
  \caption{\label{fig_kxx}
  \small 
  Bounds on $\frac{1}{d} \log \ex{ k(X,X)}$ for a distribution satisfying the sub-gamma condition with parameters $(1,d^{-\delta})$ as a function of $\delta$ for various $d$.  In all cases, $p=1$, $\sigma = 0.1$, and $k$ is the `two-moment' kernel described in Theorem~\ref{thm_kbound_subgamma}.  The upper bounds (solid line) are given by the right-hand side of \eqref{eq_ub-cor-subgamma}. The lower bounds (dashed line) are given by the right-hand side of \eqref{eq_kxx_example_LB} evaluated at $\eps = \sqrt{d}$. 
  }
\end{figure*}

The bounds on $\ex{k(X,X)}$ given in \eqref{eq_ub-cor-subgamma} and \eqref{eq_kxx_example_LB} are shown in Figure~\ref{fig_kxx} as a function of $\delta = (\log d)/(\log b)$ for various values of $d$. The plot demonstrates a phase transition at the critical value of $\delta = 0.5$.
Further computational results on this example are given in Section~\ref{subsec_exp-sub-gamma}.

\subsection{Dependent Samples}\label{sec:dependent_samples}

Motivated by applications involving Markov chain Monte Carlo there is significant interest in understanding the rate of convergence when there is dependence in the samples. Within the literature, this question is often address by focusing on stationary sequences satisfying certain mixing conditions \cite{peligrad1986recent}. The basic idea is that the dependence between $S_i$ and $S_j$  decays rapidly as $|i -j|$ increases, then the effect of the dependence is negligible. 

To go beyond the usual mixing conditions, a particularly useful property of the kernel MMD distance is that the second moment of $\gamma_k(P,P_n)$  depends only on the \emph{pairwise} correlation in the samples. This perspective is useful for settings where there may not be a natural notion of time. 

To make things precise, suppose that $S_1, \dots, S_n \in \reals^d$ is a collection of (possibly dependent) samples with identical distribution $P$. For each $i \ne j$ let $Q_{ij}$ denote the law of $(S_i, S_j)$. 
Starting with \eqref{eq_gamma_k_alt}, the expectation of the squared MMD can now be decomposed as
\begin{align}
  \bEx \left[ \gamma^2_k(P, P_n)  \right] & = \frac{1}{n}   \bEx \left[ \gamma^2_k(P, P_1)  \right]   + \frac{1}{n^2}  \sum_{i \ne  j} r_{ij}, \label{eq:gamma_k_second_moment_dependent}
\end{align}
where $r_{ij} \coloneqq \bEx_{Q_{ij}}[k(S_i, S_j)]  - \bEx_{P\otimes P}[k(S_i, S_j')]$. Notice that the first term in this decomposition is the second moment under \emph{independent} samples. The second term is non-negative and  depends only on the pairwise dependence, i.e., the difference between $Q_{ij}$ and the probability measure obtained by the product of its marginals. 

In some cases, it is natural to argue that only a small number of the terms $r_{ij}$ are nonzero. More generally it is desirable to provide guarantees in terms of measures of dependence that do not rely on the particular choice of kernel. The next result provides such a bound in terms of the Hellinger distance.

\begin{lemma}\label{lem:MMD_converge}
If $\bEx_{P}[k^2(X,X)]^{1/2} < C_{k,P}$,  then
\begin{align*}
r_{ij} \le \sqrt{2}\, C_{k,P} \,  d_H(Q_{ij}, P \otimes P ),
\end{align*}
where $d_{H}$ denotes the Hellinger distance. 
\end{lemma}

The following example is inspired by random feature kernel interpretation of neural networks ~\cite{rahimi2008random}.

\begin{figure*}[t]
\centering
    \raisebox{3pt}{
    \includegraphics[width=0.25 \textwidth]{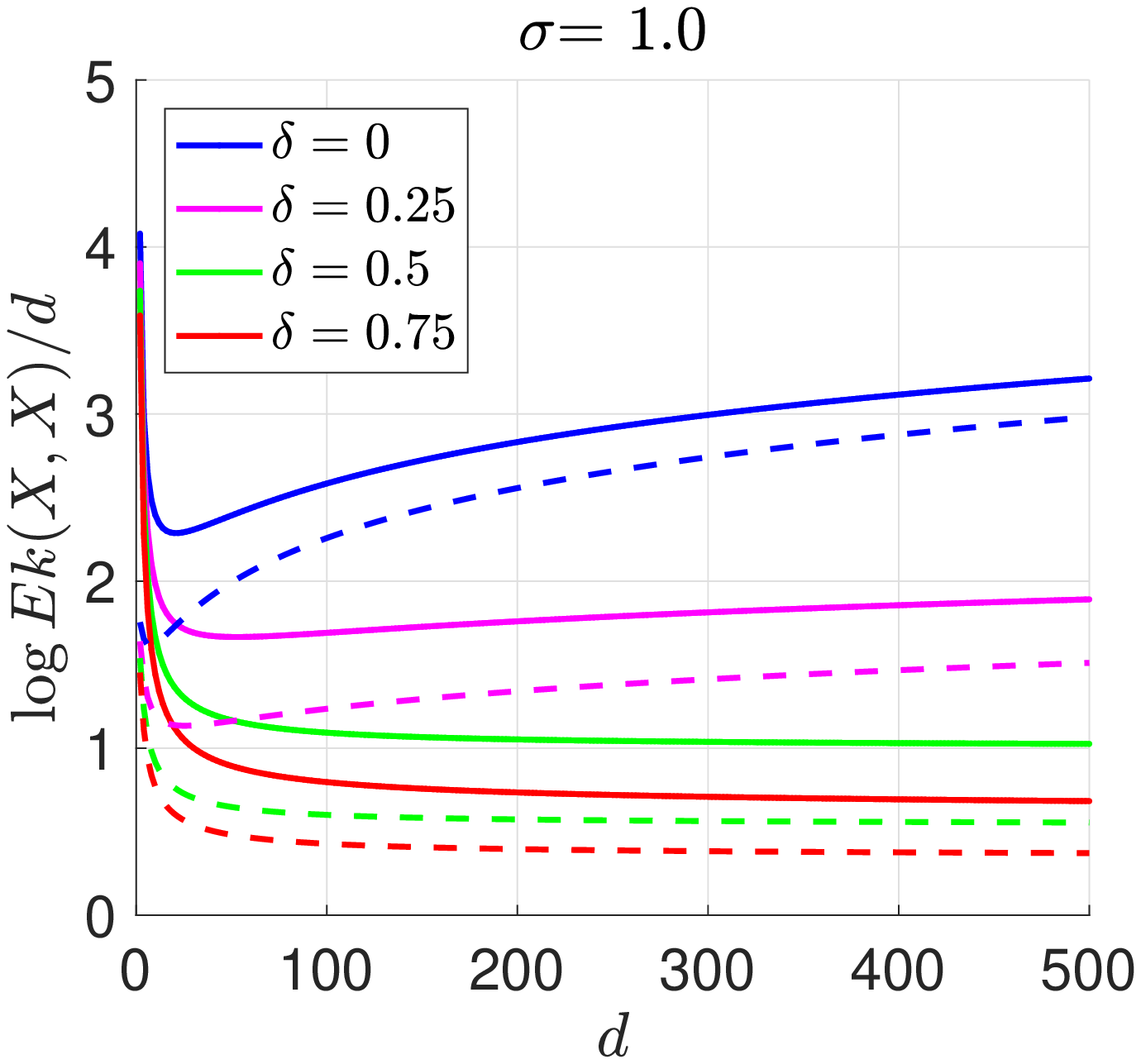}
    }
     \includegraphics[width=0.26 \textwidth]{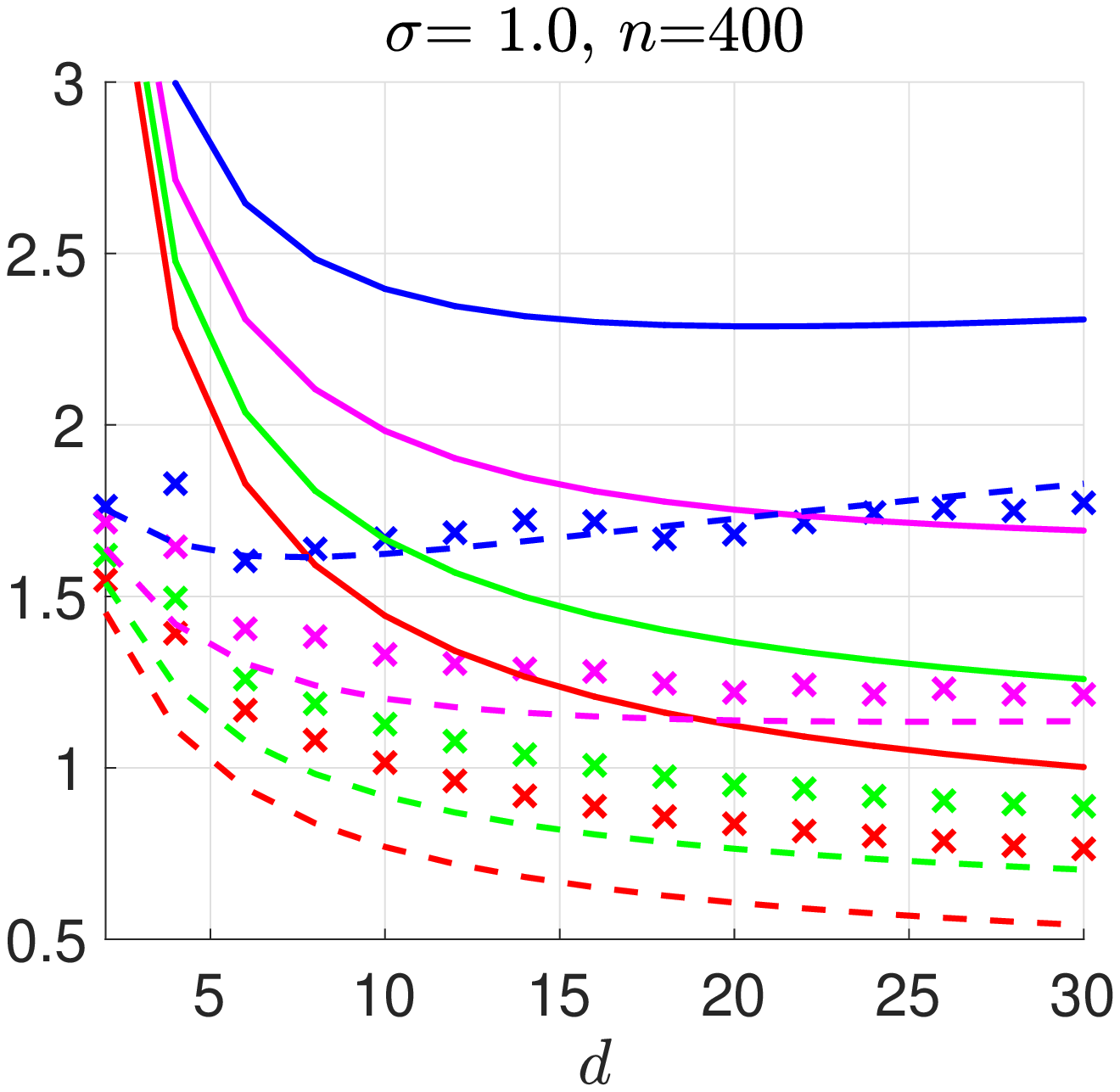}\\
        \raisebox{3pt}{
    \includegraphics[width=0.25 \textwidth]{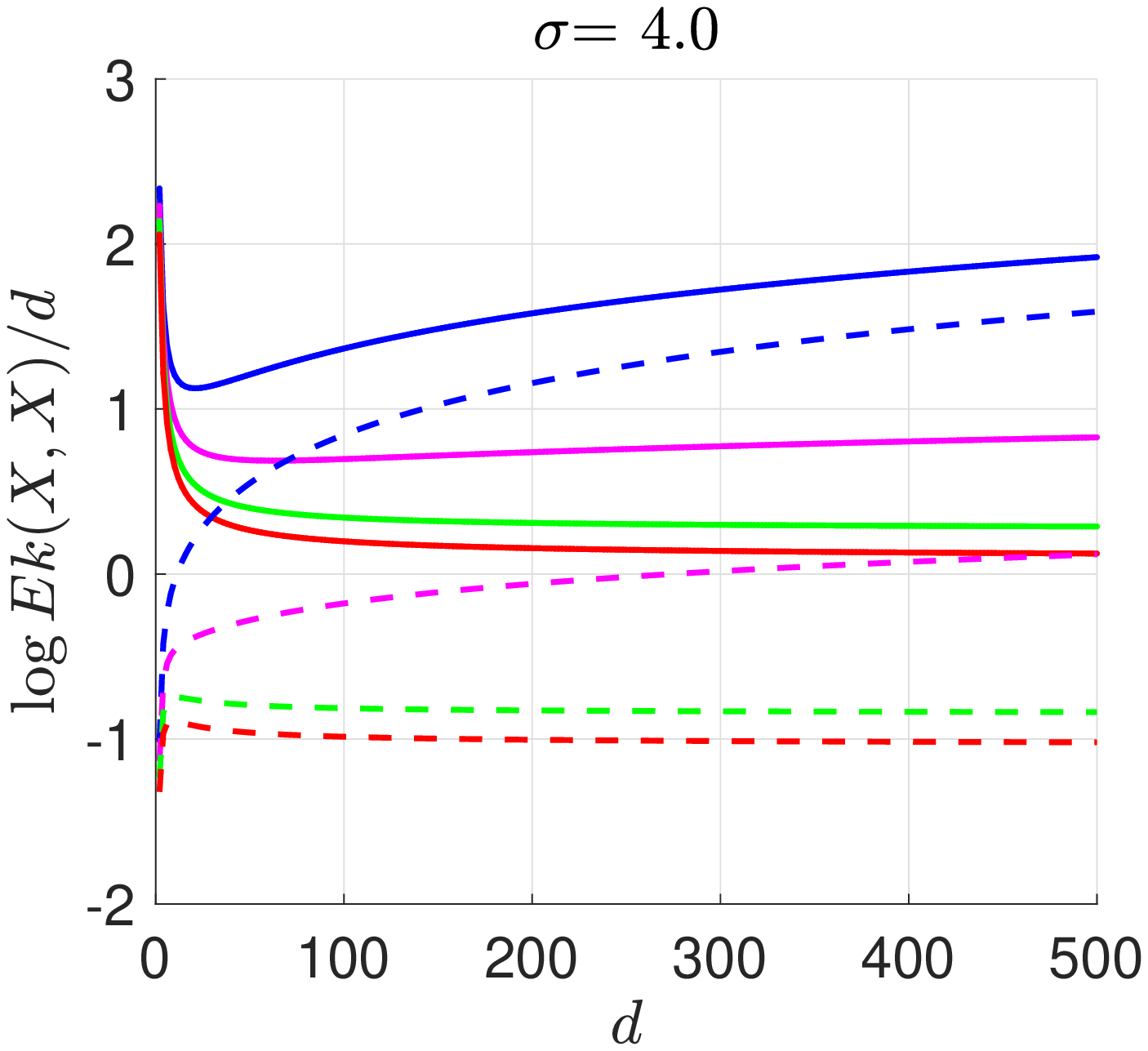}
    }
 \includegraphics[width=0.26 \textwidth]{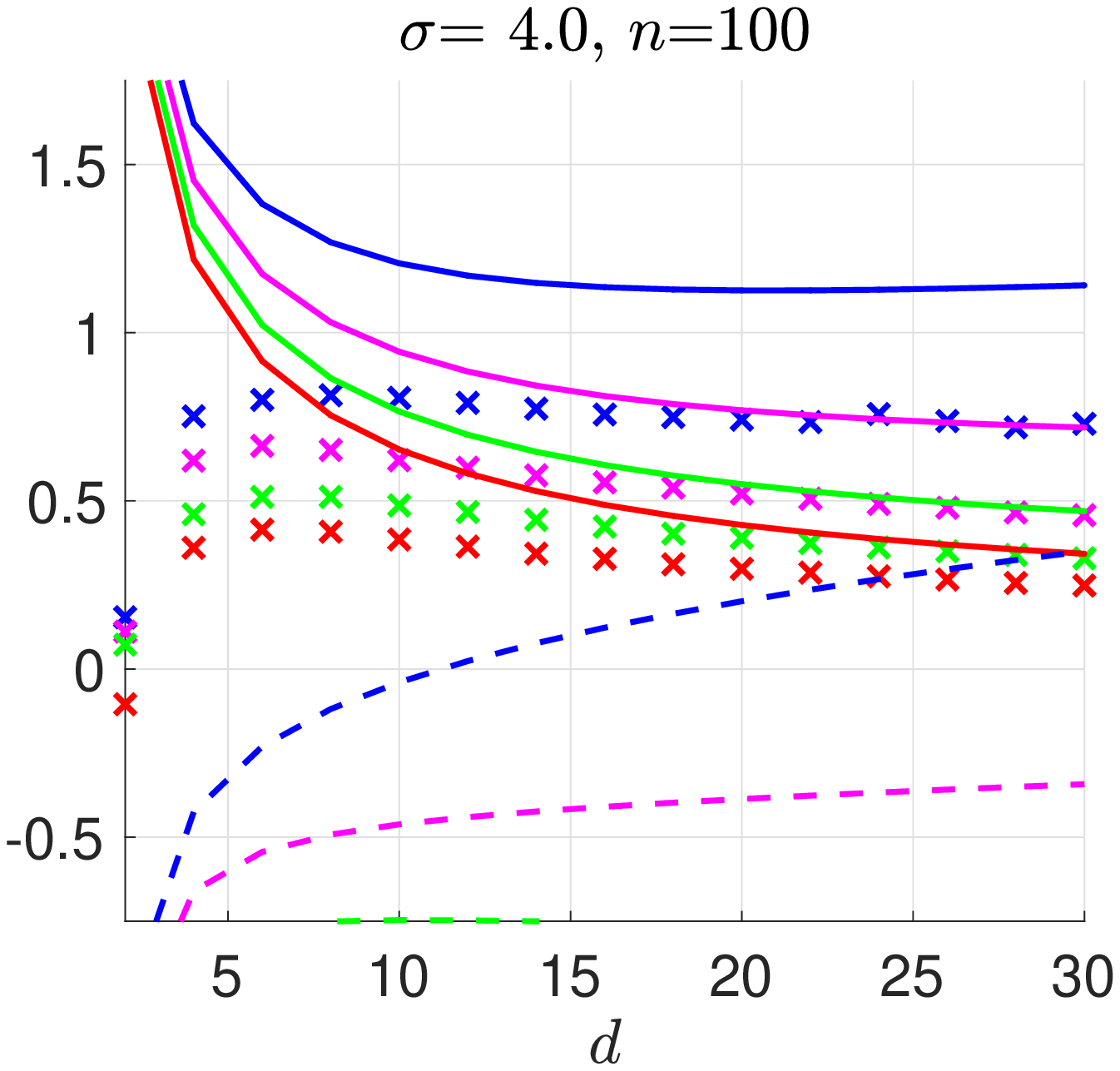}
    \caption{
    \small
    {\bf Upper panel}:
   (Left)  The UB (solid line) and LB (dashed line) of $\ex{ k(X,X)}$ given by \eqref{eq_ub-cor-subgamma}  and \eqref{eq_kxx_example_LB},
    plotted as functions of $d$ up to 500 and for various values of $\delta$ (shown in 4 colors),
    where $X$ is sub-gamma random variable in $\mathbb{R}^d$ as in Example~\ref{eg:1}, $b = d^{-\delta}$,
    and the kernel function $k$ has bandwidth $\sigma=1$.
    (Right)
    Empirical estimates of $\mathbb{E}{ \hat{\Delta}_\gamma^2 }$ (cross markers), shown together with the UB and LB  of $\ex{k(X,X)}$ as in the left plot but over a range of $d$ up to 30.
    The values are computed with  $n=400$ samples of $X$,
    and are averaged over 200 Monte-Carlo replicas.
    {\bf Bottom panel}:
    Same plots where $\sigma=4$, $n=100$, and averaged over 100 Monte-Carlo  replicas.
    In all the plots, 
    the quantities are taken $\log$ and divide by $d$ for better demonstration. 
       \label{fig:exp1}
    }
\end{figure*}

\begin{example}\label{eg:dependent-sphere}
Let $\{  Z(\alpha)\, : \, \alpha \in \reals^N \}$ be a $\reals^d$-valued Gaussian processes with mean zero and covariance function $\cov(Z(\alpha), Z(\beta)) =\langle \alpha, \beta  \rangle I_d$. Suppose that samples from $P$ are generated according to $S_i = T(  Z(\alpha_i))$ where  $\alpha_1 ,\dots, \alpha_n$ are points on the unit sphere and $T \colon \reals^d \to \reals^d$ is a function that maps a standard Gaussian vector into a vector with distribution $P$. Because Hellinger distance is non-increasing under the mapping given by $T$, it can be verified that
\begin{align*}
    d_H(Q_{ij}, P \otimes P  ) \le \sqrt{d} | \langle \alpha_i, \alpha_j\rangle|.
\end{align*}
Thus, by \eqref{eq:gamma_k_second_moment_dependent} and Lemma~\ref{lem:MMD_converge}, there exists a constant $C_k$ depending only on $k$ such that
\begin{align}\label{eq:ES-gammak2-sphere-1}
  \bEx \left[ \gamma^2_k(P, P_n)  \right] &  \le C_{k,P} \Big( \frac{1}{n} +   \frac{ 1}{n^2} \sum_{i \ne j} | \langle \alpha_i. \alpha_j\rangle|\Big).
\end{align}

This inequality holds for any set of points $\{\alpha_i\}$. To gain insight into the typical scaling behavior when the samples are nearly orthogonal on average, let us suppose that the $\{\alpha_i\}$ are drawn  independently from the uniform distribution on the sphere. Then, $\ex{ |\langle a_i, a_j \rangle|^2} = 1/N$ and by, standard concentration arguments, one finds that the following upper bound holds with high probability when $N$ is large: 
\begin{align}\label{eq:ES-gammak2-sphere-2}
  \bEx \left[ \gamma^2_k(P, P_n)  \right]  &  \le C_{k,P}\Big( \frac{1}{n} +    \sqrt{ \frac{ \log N}{N}} \Big).
\end{align}
\end{example}

\section{Numerical Results}

In this section, we compute the upper bounds of GOT distance provided by the the empirical kernel MMD distance.


\begin{figure*}[t]
\centering
   \includegraphics[width=0.32 \textwidth]{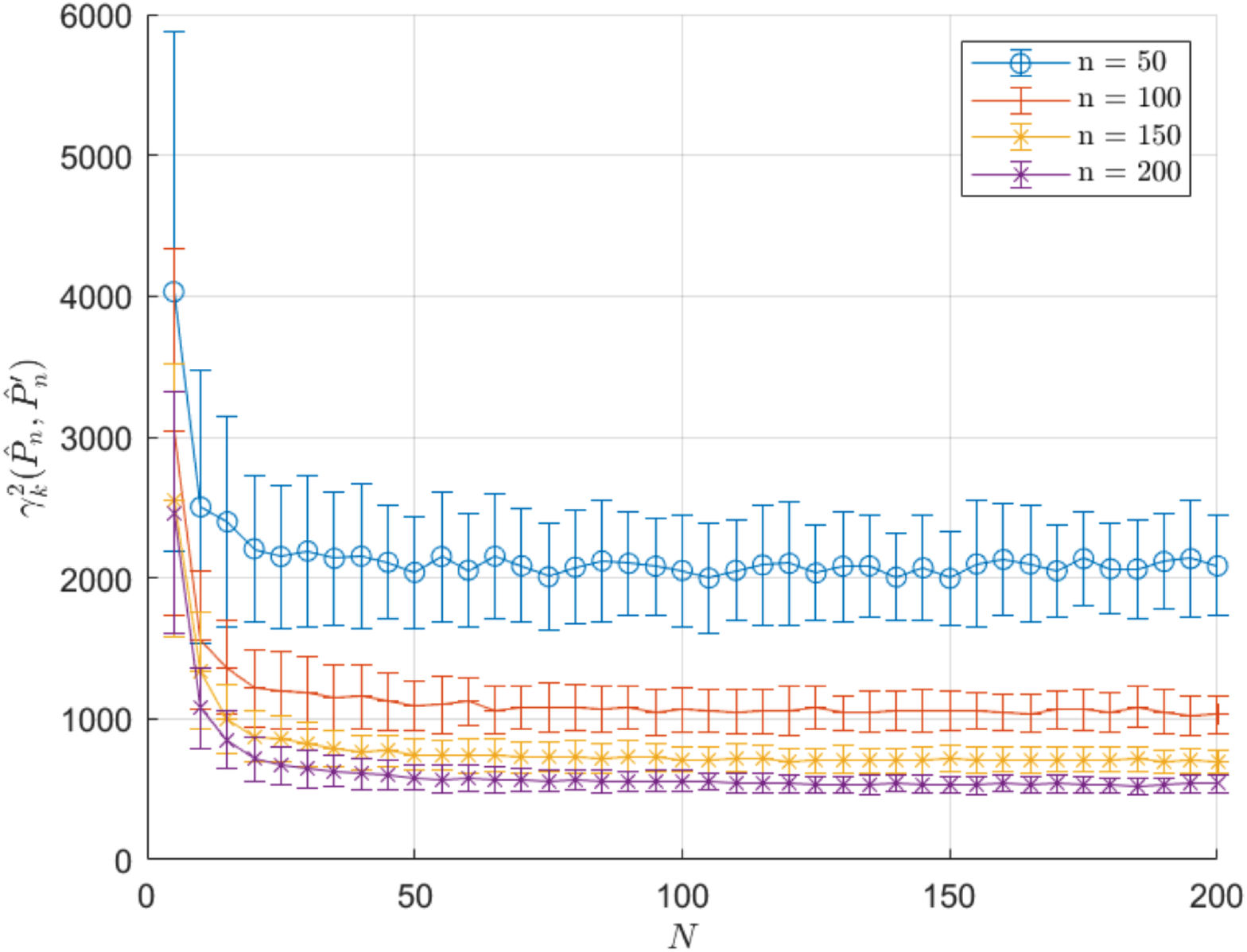}
        \includegraphics[width=0.31 \textwidth]{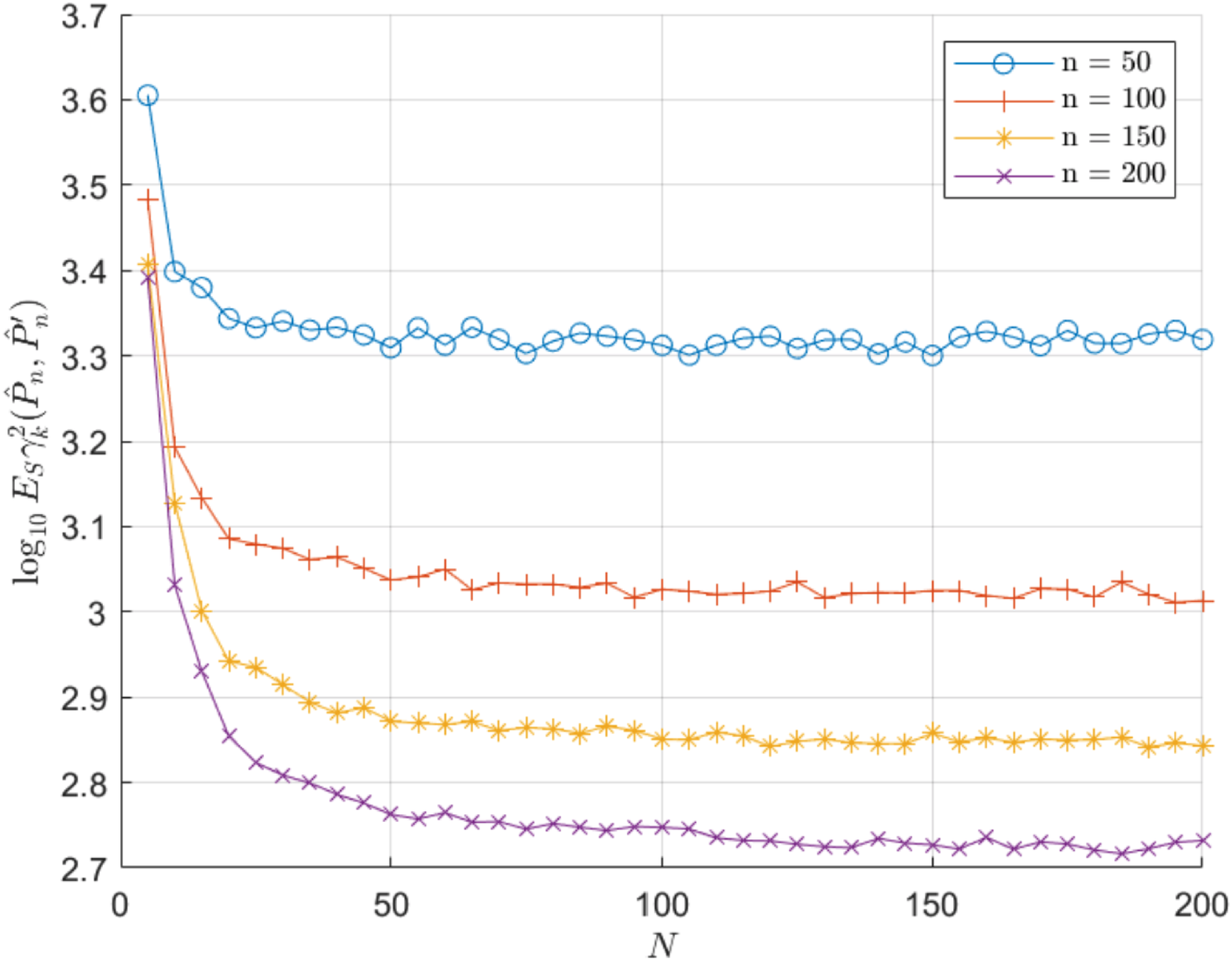}
                \includegraphics[width=0.34 \textwidth]{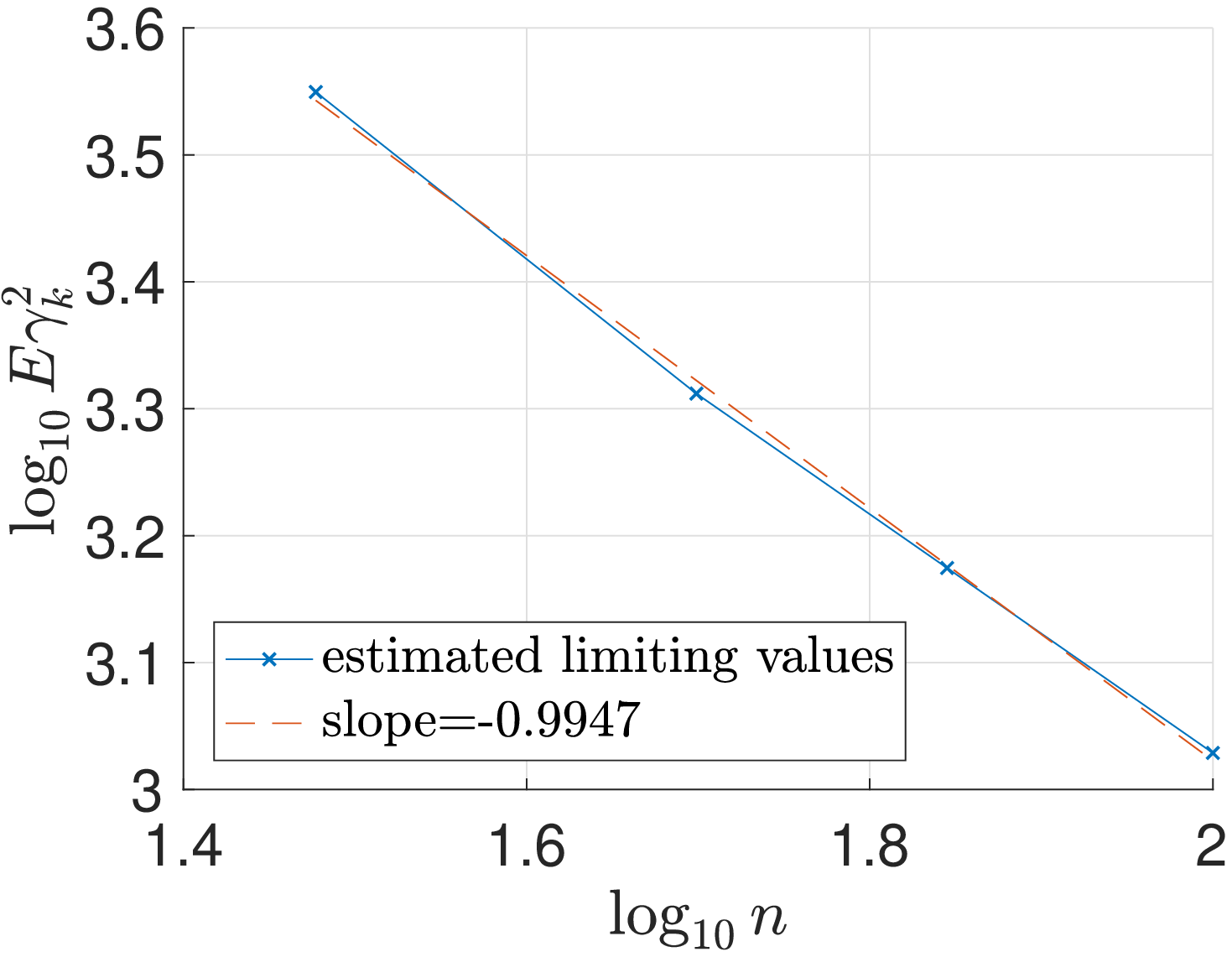}
    \caption{ \small
  Empirical values of $\gamma_k^2 (P_n, P_n')$ 
    for dependent samples in Example \ref{eg:dependent-sphere} in Section \ref{sec:dependent_samples}.
    (Left) Mean and standard deviation computed over 200 realizations of $S_i$'s,P
    plotted against the increasing values of $N$ and for different $n$.
    (Middle)  The $\log_{10}$ of the mean value in the left plot. 
    (Right) Log-log plot of the estimated limiting values of the mean in the left plot,
    by averaging over the range of $N \in [80, 100]$,
    v.s. $n$, showing a slope close to 1.
       \label{fig:exp2}
       }
       \vspace{-5pt}
\end{figure*}

\subsection{Bounds under Sub-gamma Condition}\label{subsec_exp-sub-gamma}

The sub-gamma condition allow us to address distributions that do not satisfy the  sub-Gaussian condition. Upper bounds on the convergence rate for this class of distributions follow from Theorem \ref{thm_two_moment}
combined with Theorem \ref{thm_kbound_subgamma}. 

For sub-gamma  $X$ as in  Example \ref{eg:1}, $\ex{ k(X,X)}$ has upper bound  (UB)  and lower bound (LB) as in \eqref{eq_ub-cor-subgamma}  and \eqref{eq_kxx_example_LB}  respectively.
Here, in addition to the theoretical UB and LB as shown in Figure \ref{fig_kxx},
we  also compute the estimate of 
$\ex { n \gamma_k^2( P, P_n) }$  approximated by the two-sample estimator. 
Let $P_n$ and $P_n'$ be the empirical measure defined as in \eqref{eq:def-Pn} for independent iid copies of the sub-gamma random vector as in Example \ref{eg:1},
 $\{X_i \}_{i=1}^n$ and $\{X_i' \}_{i=1}^n$,
  respectively.
Define
$\hat{\Delta}_\gamma^2 \coloneqq \frac{n}{2} \gamma_k^2( {P}_n, {P}_n')$,
and then 
$ \bEx [\hat{\Delta}_\gamma^2]  =\ex{k(X,X)} -  \ex{k(X,X')} = \ex{  n \gamma_k^2( P, P_n) }$. Because the empirical kernel MMD (squared) distance
$\gamma_k^2( P_n, P_n')$ can be computed numerically from the two samples of sub-gamma random vectors, we can estimate the   expectation of $\hat{\Delta}_\gamma^2$ by empirical average over Monte-Carlo replicas.
Detailed numerical techniques to compute the two-moment kernel and experimental setup are provided in the \appnamesmall.

The results are shown in Figure~\ref{fig:exp1}, where we set the parameter $b$ controlling the shape and scale of $X$ as $b = d^{-\delta}$, $\delta = \{ 0, 0.25, 0.5, 0.75 \}$.
The data dimension $d$ takes multiple values, and the kernel bandwidth $\sigma$ takes values 1 and 4. In both cases, the left plot shows the same information as Figure \ref{fig_kxx} in view of increasing $d$, so as to be compared to the right plot.  The right plot focuses on the case of small $d$, where the empirical estimates of $\mathbb{E} [\hat{\Delta}_\gamma^2]$ observe the UB.
Notably, these values also lie between the UB and LB (recall that the LB applies to $\ex{k(X,X)}$ but not $\mathbb{E}[ \hat{\Delta}_\gamma^2]$) for both cases of $\sigma$, and approach the LB when $\sigma=1$.
This shows that our theoretical UB captures an important component of the kernel MMD distances for this example of $X$.

\subsection{Dependent Samples via Gaussian Process}\label{sec:dependent_samples_example}

We generate dependent samples following a Gaussian process $\{ Z(\alpha)\}_{ \alpha \in S^N }$
as in Example \ref{eg:dependent-sphere} in Section \ref{sec:dependent_samples},
and numerically compute the values of $\gamma_k^2( P, P_n)$
by its two-sample version $\gamma_k^2( P_n,  {P}_n')$, using the same kernel as in the first experiment. 
Theoretically, when $n$ is large,
$\gamma_k^2(P, P_n)$ is expected to concentrate at its mean value as in \eqref{eq:ES-gammak2-sphere-1},
and then since $\alpha_i$ are uniformly sampled on the $N$-sphere,
we also expect concentration at the value of \eqref{eq:ES-gammak2-sphere-2}. 

We set $n=\{30, 50, 70, 100 \}$, and vary $N$ from 5 to 100.
The data are in dimension $d=5$, the kernel parameters are $\sigma = 0.5$, $c=1$, $p=1$, and  for each value of $n$ and $N$, $\gamma_k^2( P_n, P_n')$ are computed for 100 realizations of the dependent random variables $S_i$, conditioned on one realization of  the vectors $\alpha_i$'s. 
The results are shown in  Figure \ref{fig:exp2}, where for the small values of $N$, the computed approximate values of $\mathbb{E} \left[ \gamma_k^2(P, P_n)\right]$ are decreasing because they are dominated by the contribution from the dependence in the samples, namely the part that is upper-bounded by the second term depending on $\alpha$ in \eqref{eq:ES-gammak2-sphere-1}.
As $N$ increases, these values converge to certain positive value which decrease as $n$ increases. 

Specifically, we take the average over the mean values of $\gamma_k^2( P_n, P_n')$  for $N \in [80, 100]$ as an estimate of the limiting values, and Figure \ref{fig:exp2} (right) shows that these values decay as $n^{-1}$, as they correspond to  the first term in \eqref{eq:ES-gammak2-sphere-1}.
These numerical results show the competing two factors as predicted by the analysis in Section \ref{sec:dependent_samples}.

\section{Conclusion and Discussion}

The paper proves new convergence rates of GOT under general settings. Our results require only a finite moment condition and in the special case of sub-gamma distributions, we quantify the dependence on the dimension $d$ and show a phase transition with respect to the scale parameter. Furthermore, our results cover the setting of dependent samples where convergence is proved only requiring a condition on pairwise sample dependence expressed by the kernel. 
Throughout our analysis, the main theoretical technique is to establish an upper bound using a kernel MMD where the kernel is called a ``two-moment'' kernel due to its special properties. 
The kernel depends on the cost function of the OT as well as the Gaussian smoothing used in GOT.

For the tightness of the kernel MMD upper bound,
as has been pointed out in the comment beneath Theorem \ref{thm_GOT_moment_UB}, our result shows that the convergence rate of  $n^{-1/2}$ is tight 
in some regimes where $\sigma \to 0$ with $n \to \infty$ and the result matches the minimax rate in the unsmoothed OT setting. Alternatively, if $\sigma$ is bounded away from zero then it may be possible to obtain a better rate of convergence.
For example,  Proposition~6 in \cite{goldfeld:2020b} 
shows that if  the pair $(P,\sigma)$ satisfies an  additional chi-square divergence condition, then  $\mathcal{T}_2^{(\sigma)}(P, P_n)$ converges at rate $1/n$, which is faster than the general upper bound of $1/\sqrt{n}$ appearing in our paper. 
In this direction, pinning down the exact convergence rate in terms of regularity conditions on $P$ remains an interesting open question for future work. 
In addition, it would be interesting to investigate the relationship between the Gaussian smoothing used in this paper and the multiscale representation  of  $\cT_p$ in terms of partitions of the support of $P$,
which was used in the analysis in \cite{fournier:2015} as well as the related work \cite{weed2019sharp}.

In practice, the tightness of the kernel MMD upper bound also depends on the choice of kernel, which can be optimized for the data distribution and the level of smoothing in GOT. The question of whether the kennel MMD provides a useful alternative to OT distance in applications can be worthwhile of further investigation.  
Finally, another important direction of future work is to study computational methods and applications of the GOT approach, particularly in the high dimensional space. 
Currently, no specialized algorithm for GOP from finite samples has been developed,
except for the direct method of applying any OT algorithm, e.g., entropy OT (Sinkhorn), to data with additive Gaussian noise \cite{goldfeld:2020}.
Progress on the computational side will also enable various applications of GOT, e.g., the evaluation of generative models in machine learning.





\appendix

\setcounter{figure}{0}
\setcounter{equation}{0}
\setcounter{theorem}{0}

\ifarxiv
\renewcommand{\thefigure}{A.\arabic{figure}}
\renewcommand{\theequation}{A.\arabic{equation}}
\renewcommand{\thetheorem}{A.\arabic{theorem}}
\else 
\renewcommand{\thefigure}{S.\arabic{figure}}
\renewcommand{\theequation}{S.\arabic{equation}}
\renewcommand{\thetheorem}{S.\arabic{theorem}}
\fi

\ifarxiv
\else
\clearpage
\thispagestyle{empty}
\onecolumn
\aistatstitle{Convergence of  Gaussian-smoothed optimal transport distance with sub-gamma distributions and  dependent samples \\
Supplementary Materials}
\fi

\section{Upper Bounds on GOT} 
\label{sec_proof_UB}

This section provides proofs of the upper bounds on the GOT provided in  Section~3. For the convenience of the reader we repeat some of the necessary definitions. Recall that the feature map $\psi_x \colon \reals^d \to [0, \infty)$ is defined by
\begin{align}
    \psi_x(z) &: = \frac{ \sqrt{\omega_d}}{ 2^\frac{d+p}{2}} \frac{ \|z\|^\frac{d-1  + 2p}{2}}{ \sqrt{ f(\|z\|)}} \phi\left(\frac{ z}{\sqrt{2}}  - \frac{ x}{\sigma} \right ), \label{seq:psi}
\end{align}
where  $\omega_d = 2\pi^{d/2} / \Gamma(d/2)$ is the volume of the unit sphere in $\reals^d$, $\phi(u) = (2\pi)^{-d/2} \exp( -\frac{1}{2} \|u\|^2)$ is the standard Gaussian density on $\reals^d$, and $f$ is a probability density function on $[0, \infty)$ that satisfies 
\begin{align}
f(x) \ge a x^{d + 2p -1}  \exp( - b  x^2),
\label{eq:fLB}
\end{align}
for some $a > 0$ and $b \in (0,1/2)$.

\begin{lemma} The feature map in \eqref{seq:psi} defines a positive semidefinite kernel  $k \colon \reals^n \times \reals^n \to [0, \infty)$ according to  
\begin{align}
k(x,y) \coloneqq \int_{\reals^d} \psi_x(z) \psi_y(z) \, dz.
\end{align} 
Furthermore, this kernel can also be expressed as
\begin{align}
     k(x,y) & = \exp\left( -\frac{ \|x-y\|^2}{ 4 \sigma^2}\right) I_f\left( \frac{ \|x+y\|}{\sqrt{2} \sigma} \right), \label{seq_kernel_f}
\end{align}
where
\begin{align}
    I_f(u) &\coloneqq \frac{ \omega_d}{2^{d + p} ( 2\pi)^{\frac{d}{2}} }\int_0^\infty  \frac{ x^{d-1+2p}}{ f(x)} g_{d,u}\left(x \right)   \, dx, \label{eq:If} 
\end{align}
and $g_{d,u}(x)$ is the density of $\|Z\|$ when $Z \sim \normal(\mu, I_d)$ with $\|\mu\| = u$.
\end{lemma}
\begin{proof}
First we establish that $\psi_x$ is square integrable. By the assumed lower bound in \eqref{eq:fLB} and the fact that $\phi^2(y/\sqrt{2}) = (2 \pi)^{d/2}  \phi(y)$, we can write
\begin{align}
\int_{\reals^d}  |\psi_x(z) |^2 \, dz &\le \frac{C_{d,p} }{a}  \int_{\reals^d}  \exp\left( - b \|z\|^2 \right) \phi\Big( z  - \frac{ \sqrt{2} x}{\sigma} \Big ) \, dz.
\end{align}
This integral is the moment generating function of the non-central chi-square distribution with $d$ degrees of freedom and non-centrality parameter $2 \|x\|^2/\sigma^2$ evaluated at $b$. Under the assumption $b < 1/2$, this integral is finite. 

To establish the form given in \eqref{seq_kernel_f} we can expand the squares to obtain:
\begin{align*}
\phi\left(\frac{ z}{\sqrt{2}}  - \frac{ x}{\sigma} \right ) \phi\left(\frac{ z}{\sqrt{2}}  - \frac{ y}{\sigma} \right ) 
& = (2\pi)^{-\frac{d}{2}}  \exp\left( - \frac{\|x-y\|^2}{4 \sigma^2} \right)\phi\left(z  - \frac{ x  +y}{\sqrt{2} \sigma} \right ).
\end{align*}
Since the first factor does not depend on $z$, it follows that
\begin{align*}
     k(x,y) & =\frac{ \omega_d}{2^{d + p} ( 2\pi)^{\frac{d}{2}} }  \exp\left( -\frac{ \|x-y\|^2}{ 4 \sigma^2}\right) \int_{\reals^d} \frac{ \|z\|^{d -1+2p}}{ f(\|z\|)} \phi\left(z  - \frac{ x  +y}{\sqrt{2} \sigma} \right ) \, dz. 
\end{align*}
In this case, we recognize the integral as the expectation of $\|\cdot\|^{d-1 + 2p}/f(\cdot)$ under the chi-distribution with $d$ degrees of freedom and parameter $u = \| x + y\|/(\sqrt{2}\sigma)$. 
\end{proof}

\subsection{Proof of Theorem~2}

The following result is an immediate consequence of \cite[Theorem~6.13]{villani2008optimal} adapted to the notation of this paper. 

\begin{lemma}[{\cite[Theorem~6.13]{villani2008optimal}}]\label{lem:TVbnd}
For any $P, Q \in \cP_p(\reals^d)$, 
\begin{align}
    \cT_p(P,Q) \le 2^{\max(p-1,0)} \int \|x\|^p\, d|P - Q|(x),
\end{align}
where $|P-Q|$ denotes the absolute variation the signed measure $P-Q$.
\end{lemma}

To proceed, let $p_\sigma(z) = \int_{\reals^d} \phi_{\sigma}(z -x) \, dP(x)$ and $ q_\sigma(z) = \int_{\reals^d} \phi_{\sigma}(z -x) \, dQ(x)$ denote the probability density functions of $P \ast \normal_\sigma$ and $Q \ast \normal_\sigma$, respectively.  By Lemma~\ref{lem:TVbnd}, the OT distance between $P\ast \normal_\sigma$ and $Q \ast \normal_\sigma$ is bounded from above by the weighted total variation distance:
\begin{align}
\cT^{(\sigma)}_{p}(P,Q) \le 2^{\max(p-1,0)} \int \|z\|^p\left|  p_\sigma(z) - q_{\sigma}(z) \right| \, dz. \label{seq:TVbnd}
\end{align}

In the following we will show that $2^{\max(p-1,0)}\sigma^p \gamma_k(P,Q)$ 
provides an upper bound on the right-hand side of \eqref{seq:TVbnd}.  To proceed, recall that the kernel MMD can be expressed as
\begin{align}
\gamma^2_k(P,Q) = \ex{ k(X,X')}  + \ex{ k(Y,Y')} - 2 \ex{ k(X,Y)},  \label{seq:kernel_alt}
\end{align}
where $X,X'$ are iid $P$ and $Y,Y'$ are iid $Q$.  The assumptions  $\int \sqrt{k(x,x)} \, P(x) < \infty $ and $\int \sqrt{k(x,x)} \, Q(x) < \infty$  ensure that these expectations are finite, and so, by Fubini's theorem, we can interchange the order of integration:
\begin{align*}
\ex{ k(X,Y)} &=  \int k(x,y)\, d P(x) \, d Q(y)  = \int \left(\int \psi_x(z) \, d P(x) \right)  \left( \int \psi_x(z) \, d Q(x) \right)\, dz.
\end{align*}
For each $z \in \reals^d$, it follows that
\begin{align*}
     \int \psi_x(z) \, d P(x) & =  \frac{ \sqrt{\omega_d}}{ 2^\frac{d+p}{2}} \frac{ \|z\|^\frac{d-1  + 2p}{2}}{ \sqrt{ f(\|z\|)}}  \int_{\reals^d} \phi\left(\frac{ z}{\sqrt{2}}  - \frac{ x}{\sigma} \right ) \, dP(x) \\
& =  \frac{ \sigma^d \sqrt{\omega_d}}{ 2^\frac{d+p}{2}} \frac{ \|z\|^\frac{d-1  + 2p}{2}}{ \sqrt{ f(\|z\|)}}  p_{\sigma}\left( \frac{\sigma z}{\sqrt{2}} \right),
\end{align*}
and this leads to 
\begin{align*}
\ex{ k(X,Y)} & =  \frac{ \sigma^{2d} \omega_d}{ 2^{d+p} }   \int \frac{ \|z\|^{d -1  +2p}}{ f(\|z\|) }   p_{\sigma}\left( \frac{\sigma z}{\sqrt{2}} \right) q_{\sigma}\left( \frac{\sigma z}{\sqrt{2}} \right) \, dz \\
& =   \sigma^{-2p}    \int \frac{ \|z\|^{2p}}{ r_{\sigma}(\|z\|) }    p_\sigma(z) q_{\sigma}(z) \, dz ,
\end{align*}
where $r_{\sigma}(x) \coloneqq  \frac{ \sqrt{2}}{\sigma}  f( \frac{\sqrt{2}}{\sigma} x) / ( \omega_d \|z\|^{d-1} ) $.  Combining this expression with \eqref{seq:kernel_alt} leads to 
\begin{align*}
\gamma^2_k(P,Q) =   \sigma^{-2p} \int \frac{ \|z\|^{2p}}{ r_{\sigma}(\|z\|) }    \left( p_\sigma(z) -  q_{\sigma}(z) \right)^2 \, dz. 
\end{align*}
Finally,  we note that $z \mapsto r_{\sigma}(\|z\|)$ is a probability density function on $\reals^d$ (it is non-negative and integrates to one) and so by Jensen's inequality and the convexity of the square, 
\begin{align*}
\gamma^2_k(P,Q) & \ge   \sigma^{-2p}  \left(  \int  \|z\|^p    \left| p_\sigma(z) -  q_{\sigma}(z) \right| \, dz \right)^2 .
\end{align*}
In view of \eqref{seq:TVbnd}, this establishes the desired result. 

\subsection{Proof of Theorem~3}
The fact that the kernel MMD provides an upper bound on $\cT_p^{(\sigma)}(P,Q)$ follows directly from Theorem~2. All the remains to be shown is that $\sqrt{k(x,x)}$ is integrable for any probability measure with finite $s$-th moment, where $s = (d + 2p + \eps)/2$. To this end, we note that by the triangle inequality, 
\begin{align*}
    M_{d,u}(r) \le  2^{\min(1,r)}  \left( M_{d}(r)   + \|u\|^r \right),
\end{align*}
for all $r \ge 0$. Under the assumptions on $\eps$, we have $0 \le d + 2p - \eps < d + 2p + \eps \le 2s$ and so there exists a constant $C_{d,p,\eps,\lambda}$ such that 
\begin{align*}
k(x,y) \le C_{d,p,\eps,\lambda} \left(1 + \|x\|^{2s} + \|y\|^{2s}  \right) .
\end{align*}
Thus, the existence of finite $s$-th moment is sufficient to ensure that $\sqrt{k(x,x)}$ is integrable. 

\section{Convergence Rate} 

This section provides proofs for the results in Section 4 of the main text as well as Theorem~1.  To simplify the notation, we define  $r = d + 2p$ and let $Y = (\sqrt{2}/\sigma) X + Z$ where $Z  \sim \normal(0, I_d)$.

Let us first consider some properties of $\ex{k(X,X)}$. Since the two moment kernel satisfies $k(x,x) = \alpha_{d,p} \, J( ( \sqrt{2} / \sigma ) \|x\| )$, it follows from the definition of $J(\cdot)$ that
\begin{align}
\ex{ k(X,X) } = \frac{ \alpha_{d,p}  }{ 2 \eps} \left( \lambda^\eps\, \ex{ \| Y\|^{r -\eps}}  + \lambda^{-\eps} \,\ex{ \|Y\|^{r + \eps}}   \right) . \label{eq:kxx_exact}
\end{align}
Suppose that there exists numbers $M_-$ and $M_+$ such that
\begin{align}
\ex{ \|Y\|^{r - \eps} } \le M_-  , \quad \ex{ \|Y\|^{r  + \eps} } \le M_+. \label{seq:Yr_bnd}
\end{align}
Choosing 
\begin{align}
\lambda = ( M_+/M_-)^{1/(2 \eps)}, \label{sec:lambda_opt}
\end{align}
leads to 
\begin{align}
\ex{ k(X,X) } \le  \frac{ \alpha_{d,p}  }{ \eps}  \sqrt{M_- M_{+}}. \label{seq:kxx_UB}
\end{align}
In other words, optimizing the choice of $\lambda$  results in an upper bound  on $\ex{k(X,X)}$ that depends on only the geometric mean  of the upper bounds on  $ \ex{ \|Y\|^{r \pm \eps}} $.

\begin{lemma}\label{lem:Mpm_bnd}
Let $X \in \reals^d$ be a random vector satisfying
\begin{align}
   \left( \ex{ \|X\|^{s}} \right)^\frac{1}{s}  \le m(s),
\end{align}
for some function $m(s)$ for $s \ge 1$.  Then, if $r -\eps \ge 1$,  \eqref{seq:Yr_bnd} holds with
\begin{align}
    M_{\pm} & = \left( \left(\overline{M}_{d}(r \pm \eps)\right)^{\frac{1}{r \pm \eps}}  + \frac{ \sqrt{2}}{\sigma}  m(r \pm \eps)  \right)^{r \pm \eps},
\end{align}
where $\overline{M}_d(s) = (d+s)^{(d+s -1)/2}d^{-(d-1)/2} e^{-s/2}$. 
\end{lemma}
\begin{proof}
This result follows from Minkowski's inequality, which gives
\begin{align*}
    \left( \ex{ \|Y\|^s} \right)^\frac{1}{s}  & \le  \left( \ex{ \|Z\|^{s} }\right)^\frac{1}{s}   + \frac{ \sqrt{2}}{\sigma} \left( \ex{ \|X\|^{s} }\right)^\frac{1}{s}
\end{align*}
for all $s \ge 1$ and the upper bound on $M_{d}(s) = \ex{ \|Z\|^s}$ in Theorem~5. 
\end{proof}

\subsection{Proof of Theorem~1} 
The result follows immediately by combining Theorem 4 and Equation (11) in the main text.

\subsection{Proof of Theorem~4} 
By Lyapunov's inequality  and Minkowski's inequality, it follows that for  $t \in \{ r \pm \eps\}$, 
\begin{align*}
\left( \ex{ \|Y\|^t} \right)^\frac{1}{t}
& \le  \left( \ex{ \|Y\|^{r+\eps} } \right)^\frac{1}{r+\eps} \\
&  \le 
   \left( \ex{\|Z\|^{r+\eps}} \right)^\frac{1}{r + \eps}  +   \frac{\sqrt{2}}{\sigma}  \left( \ex{\|X\|^{r+\eps}} \right)^\frac{1}{r + \eps}   \\
  & \le \sqrt{d +  r + \eps}  + \frac{ \sqrt{2} m }{\sigma} , 
\end{align*}
where the last step holds because $M_d(q) \le (d + q)^{q/2}$ and the assumption $\left(\ex{ \|X\|^s}\right)^{\frac{1}{s}} \le m$.  Thus, for $\lambda =  \sqrt{ r + \eps}  + m $, the bound in \eqref{seq:kxx_UB} becomes
\begin{align*}
\ex{ k(X,X) } \le  \frac{ \alpha_{d,p}  }{ \eps} \left( \sqrt{d  + r + \eps}  + \frac{ \sqrt{2} m }{\sigma}   \right)^{r}.
\end{align*}
Recalling that $r = d + 2p$ gives the stated result.

\subsection{Proof of Theorem 5} 

\begin{lemma}\label{lem:subgamma_bnd}
Let $X \in \reals^d$ be a sub-gamma random vector with parameters $(v, b)$. For all $s \in [0, \infty)$ and $\lambda \in (0,1/b)$, 
\begin{align}
\ex{ \|X\|^s}  
& \le  \frac{2  \sqrt{ \pi} }{2^\frac{s}{2} \Gamma( \frac{s+1}{2} )  } \left(\frac{s}{\lambda  e} \right)^s \exp\left(  \frac{ \lambda^2 v}{ 2 (1 - \lambda b)}  \right)  M_d(s) ,  \label{eq:subgamma_UB1alt}
\end{align} 
where $M_d(s)  \coloneqq   2^\frac{s}{2}  \Gamma( \frac{d+ s}{ 2}) /  \Gamma( \frac{d}{2})$. In particular, if $\lambda =(\sqrt{ (sb)^2 + 4 vp}  -sb )/(2v) $, 
then
\begin{align}
\ex{ \|X\|^s} 
& \le  \left(  \sqrt{v +  \left(\frac{\sqrt{s} b}{2 } \right )^2  } + \frac{\sqrt{s} b}{2}  \right)^s  \frac{2  \sqrt{ \pi} }{2^\frac{s}{2} \Gamma( \frac{s+1}{2} ) } \left(\frac{s}{  e} \right)^\frac{s}{2} M_d(s)  .\label{eq:subgamma_UB2}
\end{align} 
\end{lemma}
\begin{proof}
Let  $Y = Z^\top X$ where  $Z = (Z_1, \dots, Z_d)$ is independent of $X$ and  distributed uniformly on the unit sphere in $\reals^d$. Since $Z$ is orthogonally invariant, it may be assumed that $X = (\|X\|, 0, \dots, 0)$ and thus    $Y$ is equal in distribution to $Z_1 \|X\|$. Therefore, 
\begin{align*}
\ex{ |Y|^s} = \ex{ |Z_1|^s} \, \ex{  \|X\|^s} .
\end{align*}
The variable  $Z_1$ has density function
\begin{align*}
f(z) = \frac{\Gamma(\frac{d}{2}) }{ \sqrt{\pi}\Gamma(\frac{d-1}{2}) }(1- z^2)^{(d-3)/2}, \quad z \in [-1,1],
\end{align*}
and so the moments are given by 
\begin{align*}
\ex{ |Z_1|^s} & = \frac{2 \Gamma(\frac{d}{2}) }{ \sqrt{\pi}\Gamma(\frac{d-1}{2}) } \int_{0}^1 z^s (1- z^2)^{(d-3)/2} \, d z 
=\frac{ \Gamma(\frac{d}{2}) \Gamma(\frac{s+1}{2}) }{ \sqrt{\pi}\Gamma(\frac{d+s}{2}) } .
\end{align*}

To bound the absolute moments of $Y$ we use the basic inequality $u \le \exp(u - 1)$ with $u  =\lambda |y|/s$, which leads to
\begin{align*}
|y|^s   \le  \left( \frac{s}{\lambda e} \right)^s \exp(  \lambda|y|)  \le \left( \frac{s}{\lambda e} \right)^s ( e^{ \lambda y} + e^{- \lambda y}),
\end{align*} 
 for all $s,\lambda \in (0, \infty)$. Noting that $Y$ is equal in distribution to $-Y$ and then using the sub-gamma assumption along with the fact that $Z$ is a unit vector yields
\begin{align*}
\ex{ |Y|^s} & \le  2\left( \frac{s}{\lambda e} \right)^s \ex{ \exp( \lambda Y) } \\
& =  2\left( \frac{s}{\lambda e} \right)^s \ex{ \exp( \lambda Z^\top X ) } \\
& \le  2\left( \frac{s}{\lambda e} \right)^s  \exp\left( \frac{ \lambda^2 v}{2(1 - \lambda b)} \right).
\end{align*}
Combining the above displays yields \eqref{eq:subgamma_UB1alt}.

Finally, under the specified value of $\lambda$ it follows that 
\begin{align*}
\frac{\lambda^2 v}{1- \lambda b} = p , \qquad \frac{\sqrt{s}}{\lambda}  & =   \sqrt{v +  \left(\frac{\sqrt{s} v}{2 } \right )^2  } + \frac{\sqrt{s} b}{2} 
\end{align*}
and plugging this expression back into the bound gives \eqref{eq:subgamma_UB2}. 
\end{proof}

Theorem~5 now follows as a corollary of Lemma~\ref{lem:subgamma_bnd}. Starting with \eqref{eq:subgamma_UB2} and using the basic inequality $\sqrt{ a^2 + b^2} \le a + b$ leads to  
\begin{align*}
\ex{ \|X\|^s}& \le
\left(  \sqrt{v}   + \sqrt{s} b  \right)^s \frac{2  \sqrt{ \pi} }{2^\frac{s}{2} \Gamma( \frac{s+1}{2} ) } \left(\frac{s}{  e} \right)^\frac{s}{2} M_d(s).
\end{align*} 
To simplify the expressions involving the Gamma functions we use  the lower bound $\log \Gamma(z) \ge (z -\frac{1}{2}) \log z  - z + \frac{1}{2} \log(2 \pi)$ for $z >0$, which leads to 
\begin{align*}
 \frac{2  \sqrt{ \pi} }{2^\frac{s}{2} \Gamma( \frac{s+1}{2} ) } \left(\frac{s}{  e} \right)^\frac{s}{2} 
 &\le \sqrt{2e}     \left( \frac{s}{s+1} \right)^\frac{s}{2}.
\end{align*}
Combining this bound with the expression above yields
\begin{align*}
\ex{ \|X\|^s}  & \le \sqrt{ 2 e}   \left(  \sqrt{v}   + \sqrt{s} b  \right)^s \left( \frac{s}{s + 1} \right)^\frac{s}{2} M_d(s)\\
 & \le  \sqrt{2 e }  \left( \sqrt{v}   + \sqrt{s} b  \right)^s    M_d(s).
  \end{align*} 
This completes the proof of Theorem~5.

\subsection{Proof of Theorem~6} 

Since $Z \sim \normal(0,I_d)$ is sub-gamma with parameters $(1,0)$ it follows that  $Y =( \sqrt{2}/\sigma) X + Z$ is sub-gamma with parameters $(1 + 2 v/ \sigma^2, \sqrt{2} b/\sigma)$. For $t > -r$ we can apply   Theorem~5 to obtain
\begin{align*}
    \ex{ \|Y\|^{r+t}} \le \sqrt{2 e} \left(  \sqrt{1 + 2 v/\sigma^2} + \sqrt{r + t} \sqrt{2} b/\sigma \right)^{r +t}  \overline{M}_d(r +t) = \frac{\sqrt{2 e}}{\sigma^{r+t}}m(r+t) .
\end{align*}
Under the specified value of $\lambda = (m(\eps)/m(-\eps))^{1/(2 \eps)}$, it then follows from \eqref{seq:kxx_UB} that
\begin{align}
\ex{ k(X,X) } \le  \frac{ \sqrt{2 e}  \alpha_{k,p}  }{ \sigma^r \eps}   \sqrt{m(-\eps)  m(\eps) }. \label{seq:kxx_meps}
\end{align}

To proceed, let $(v',b') = ( \sigma^2 + 2 v,  \sqrt{2}b)$ and consider the decomposition
\begin{align*}
 \log (m(-\eps)  m(\eps))  & = 2 \log m(0) + A + B ,
\end{align*}
where
\begin{align*}
A&\coloneqq (r - \eps)  \log    \left(  \sqrt{ v'} + \sqrt{r - \eps} \,  b'  \right) +  (r + \eps) \log   \left(  \sqrt{ v'} + \sqrt{r + \eps} \,  b'  \right) - 2r  \log( \sqrt{v'} + \sqrt{r} b')\\
B &\coloneqq  \log\overline{M}_{d}(r - \eps)   + \log \overline{M}_{d}(r + \eps)  - 2 \log \overline{M}_d(r).
\end{align*} 
Using the basic inequalities 
$\sqrt{1 + x} - 1 \le x/2$ and $\log(1+x) \le x$, the  term $A$ can be bounded from above as follows:
\begin{align*}
A & = (r - \eps)  \log    \left( 1 + \frac{( \sqrt{r - \eps}   -\sqrt{r})  b'}{ \sqrt{v'} + \sqrt{r} b'}   \right) +  (r + \eps)  \log    \left( 1 + \frac{ ( \sqrt{r  +  \eps}   -\sqrt{r} ) b'}{ \sqrt{v'} + \sqrt{r} b'}   \right)\\
& \le(r - \eps)  \log    \left( 1 + \frac{( \sqrt{r - \eps}   -\sqrt{r})  }{ \sqrt{r} }   \right) +  (r + \eps)  \log    \left( 1 + \frac{ ( \sqrt{r  +  \eps}   -\sqrt{r}  }{  \sqrt{r} }   \right)\\
& \le(r - \eps)  \log    \left( 1 - \frac{\eps  }{2r}   \right) +  (r + \eps)  \log    \left( 1 + \frac{ \eps  }{ 2r}   \right)\\
& \le - (r - \eps) \frac{\eps  }{2r}  +  (r + \eps)   \frac{ \eps  }{ 2r}  \\
&= \frac{ \eps^2}{r}.
\end{align*}
Similarly, one finds that
\begin{align*}
B &=  \frac{d + r - \eps -1}{2} \log\left( 1 - \frac{\eps}{ d + r} \right) +\frac{d + r + \eps -1}{2} \log\left( 1 + \frac{\eps}{ d + r} \right) \le \frac{ \eps^2}{ d + r} .
\end{align*}
Combining these bounds with the fact that $r \ge d$ leads to 
\begin{align*}
 \sqrt{m(-\eps)  m(\eps)}   & \le   ( \sqrt{v'} + \sqrt{r} b')^r \overline{M}_d(r) \exp\left(  \frac{ 3 \eps^2}{ 4 d}  \right) .
\end{align*} 
Plugging this inequality back into \eqref{seq:kxx_meps} yields
\begin{align}
\ex{ k(X,X) } \le  \frac{ \sqrt{2 e}  \alpha_{k,p}  }{ \eps}    ( \sqrt{1 + 2 v/ \sigma^2} + \sqrt{2r} b)^r \overline{M}_d(r) \exp\left(  \frac{ 3 \eps^2}{ 4 d}  \right). \label{seq:kxx_meps_b}
\end{align}

Finally, by the lower bound  $\log \Gamma(z)\ge  (z -\frac{1}{2}) \log z  - z + \frac{1}{2} \log(2 \pi) $  and the basic inequality $(1 + p/d)^d \le e^p$ for $p,d, \ge 0$  we can write 
\begin{align*}
\alpha_{d,p} \overline{M}_d(r) 
& \le    \frac{\sqrt{ \pi }   (d+p )^{d+p} }{e^p d^{d}}   \frac{ d}{ \sqrt{ d + p}}\le    \sqrt{ \pi }   (d+p )^{p}   \sqrt{d}.
\end{align*}
Hence, 
\begin{align*}
 \ex{ k(X,X)}  \le  \sqrt{2 \pi e}   (d+p )^{p}    \sqrt{d}  \left( \sqrt{1+2v/\sigma} + \sqrt{2r} b/\sigma \right)^r  \frac{  \exp\left(  \frac{ 3 \eps^2}{ 4 d}  \right)} {\eps}.
\end{align*}
This bound holds for all $\eps \in [0,r]$. Evaluating with $\eps = \sqrt{d}$ and recalling that $r = d + 2p$ gives the stated result.

\subsection{Proof of Lemma~7}
Note that $Q_{ij}$ is absolutely continuous with respect to $P \otimes P$ and let $\lambda_{ij}= d Q_{ij}/d (P \otimes P)$ denote the Radon-Nikodym derivative. Then
\begin{align*}
    r_{ij} &= \int k (\lambda_{ij} - 1) d (P\otimes P) \\
    & =\int k (\sqrt{\lambda_{ij}} + 1)(\sqrt{\lambda_{ij}} - 1)  d (P\otimes P)\\
    & \le  \sqrt{2} \left( \sqrt{ \int k^2  d Q_{ij} } +\sqrt{ \int k^2  d (P \otimes P)} \right)  d_H(Q_{ij} , P \otimes P) , 
\end{align*}
where the last step is by the Cauchy-Scharz inequality and we have used that fact that $d^2_H(Q_{ij}, P\otimes P) =  \frac{1}{2} \int (\sqrt{\lambda} - 1)^2 d(P \otimes P)$.

Next, since $k^2(x,y) \le k(x,x) k(y,y)$ for any positive semidefinite kernel, it follows that
\begin{align*}
    \int k^2(x,y)   d Q_{ij}(x,y) \le   \int k(x,x) k(y,y)    d Q_{ij}(x,y) = \int k^2(x,x) dP(x),
\end{align*}
and thus the stated result follows from the assumption $\bEx_{P}\left[k^2(X,X)\right] \le C_{k,P}^2$. 

\section{Experimental Details and Additional Results}
In this section, we provide details of the experiments in Section 5 of the main text and additional numerical results.  Our experiments are based on the two-moment kernel given in Definition~\ref{def:two_moment}.

\subsection{Numerical Computation of the Two-Moment Kernel}\label{sec_kernel_numerical_computation}
To evaluate the two-moment kernel given in Definition~\ref{def:two_moment} we need to numerically compute the function $M_{d,u}(s)$, which is the $s$-th moment of the non-central chi-distribution with $d$ degrees of freedom and parameter $u$.  For all $s \ge 0$, this function can be written as a Poisson mixture of the (central) moments according to 
\begin{align}\label{eq_app_Mdu}
M_{d,u}(s)  & =\sum _{k=0}^{\infty }\frac {u^{2k} \exp(  - \frac{ 1}{2} u^2) }{2^k k!} M_{d+2k,0}( s).
\end{align}
This series can be approximated efficiently by retaining only the terms with $k \approx u^2 /2 $. 

Alternatively, if $s = 2\ell$ where $\ell$ is an integer, then $M_{d,u}(2\ell)$ is the $\ell$-th moment of the chi-square distribution with $d$ degrees of freedom and non-centrality parameter $u^2$. The integer moments of this distribution can be obtained by differentiating the moment generating function. An explicit formula is given by \cite[pg.~448]{johnson1995continuous}
\begin{align}
M_{d,u}(2\ell) = 2^{\ell} \Gamma( \ell + d/2)  \sum_{j=0}^\ell \binom{k}{j} \frac{ (u^2/2)^j}{ \Gamma( j + d/2)}.
\end{align}
Here we see that $M_{d,u}(2\ell)$ is a degree $\ell$ polynomial in $u^2$.

Accordingly, for any tuple $(d,p,\sigma, \lambda, \eps)$ such that $d +2p \pm \eps$ are even integers, the two-moment kernel defined in \eqref{eq:kernel_two_moment} can be expressed as
\begin{align}
    k(x,y)   = \exp\left( - \frac{ \|x-y\|^2}{4 \sigma^2} \right) \sum_{\ell=0}^{L}  c_\ell \left( \frac{\|x + y\|}{\sqrt{2}\sigma} \right)^{2\ell},
\end{align}
where $L  = (d + 2p + \eps)/2$ and the coefficients $c_0, \dots, c_{L}$ are given by
\begin{align}
    c_\ell \coloneqq  \frac{\alpha_{d,p}}{\eps 2^\ell \Gamma(\ell  + d/2)}  \left[ \lambda^\eps 2^{L-\eps} \Gamma( L - \eps  + d/2) \binom{L - \eps}{\ell} \boldsymbol{1}_{\{\ell \le L - \eps\}}   + \lambda^{\eps} 2^{L} \Gamma( L + d/2) \binom{L}{\ell} \right],
\end{align}
with $\alpha_{d,p} \coloneqq  (2 \pi) 2^{-(p+d)} 2^{-d/2}/\Gamma(d/2)$.

\subsection{Details for Example~\ref{eg:1}}\label{sec:example_sub_gamma}

We now consider Example~\ref{eg:1}, a specific example of a sub-gamma distribution which shows that the upper bound in Theorem~\ref{thm_kbound_subgamma} is tight with respect to the scaling of the dimension $d$ and the scale parameter $b$. Specifically, let $X = \sqrt{U}Z$ where $Z \sim \normal(0, I_d)$ is a standard Gaussian vector and $U$ is an independent Gamma random variable with shape parameter $1/(2 b^2)$ and scale parameter $2 b^2$. 

\begin{lemma}
For  $\alpha \in \reals^d$ such that $\|\alpha\| \le 1/b$, it holds that
\begin{align}
    \ex{\exp( \alpha^\top X)}  = - \frac{1}{2 b^2} \log\left( 1 - \|\alpha\|^2 b^2  \right). \label{eq:mgfX}
\end{align}
In particular, this means that $X$ is a sub-gamma random vector with parameters $(1, b)$. Furthermore, for $s > \max\{ -b^{-2}, -d\}$,
\begin{align}
\ex{ \|X\|^s} & =  b^s M_{b^{-2}}(s) M_{d}(s).
\end{align}
\end{lemma}
\begin{proof}
Observe that $\alpha^\top X = \sqrt{U} \alpha^\top Z$ where $\alpha^\top Z \sim \normal(0,\|\alpha\|^2)$. Hence
\begin{align*}
    \ex{\exp( \alpha^\top X)}  = \ex{ \exp\left( \frac{\|\alpha\|^2}{2} U \right)}.
\end{align*}
Recognizing the right-hand side as the moment generating function of the Gamma distribution evaluated at $\|\alpha\|^2/2$ yields \eqref{eq:mgfX}. To see that this distribution satisfies the sub-gamma condition, we use the basic inequality $
- \log(1-x) \le x/(1-x)  \le x/(1 - \sqrt{x})$ for all $x \in (0,1)$. 

The expression for the moments follows immediately from the independence of $U$ and $Z$ and the fact that $U^2/b^{2}$ has Gamma distribution with shape parameter $b^{-2}/2$ and scale parameter $2$, which implies that $U/b$ has a chi distribution with $b^{-2}$ degrees of freedom.  
\end{proof}

Since $X$ satisfies the sub-gamma condition with parameters $(1,b)$  the upper bound in Theorem~\ref{thm_kbound_subgamma} applies. Alternatively, for each pair $(\eps, \lambda)$ we can consider the exact expression for $\ex{k(X,X)}$ given in \eqref{eq:kxx_exact} 
where $r= d + 2p$ and
\[
Y = \left( \frac{2}{\sigma^2} U + 1\right)^{1/2} Z.
\]
Minimizing this expression with respect to $\lambda$ yields
\begin{align}
 \ex{ k(X,X) } \ge \frac{ \alpha_{d,p}  }{ \eps} \left( \ex{ \| Y\|^{r -\eps} } \ex{ \|Y\|^{r + \eps} }  \right)^{1/2} .
 \end{align}
To get a lower bound on the moments, we use
\begin{align}
    \ex{ \|Y\|^s} & \ge \left(\frac{\sqrt{2}}{\sigma} \right)^s \ex{ \|X\|^s} .
\end{align}
Combining the above displays leads to \eqref{eq_kxx_example_LB}. 
Using Stirling's approximation $\log \Gamma(z) = (z -\frac{1}{2}) \log z  - z + \frac{1}{2} \log(2 \pi) + o(1)$ as $z \to \infty$ it can be verified that  the minimum of this lower bound with respect to $\eps$ satisfies the same scaling behavior with respect to $d$ as the upper bound in Theorem~\ref{thm_kbound_subgamma}. Namely, the bound exponential in $d$ if $\delta \ge 1/2$ and superexponential in $d$ if $\delta < 1/2$.

\subsection{Experiments in Section~\ref{subsec_exp-sub-gamma}}

In this experiment, $p=1$, the random variable $X \in \reals^d$ is generated according to the distribution in Example \ref{eg:1},
and  the kernel bandwidth $\sigma$ takes values  $1$ and $4$. The parameters $(\lambda, \eps)$ of the two-moment kernel are specified as in Theorem~\ref{thm_kbound_subgamma} with parameters $(1,b)$, and $k(x,y)$ can be computed as in Appendix \ref{sec_kernel_numerical_computation}.

In the Monte-Carlo computation of the average of  $ \hat{\Delta}_\gamma^2$ (the right column of Figure \ref{fig:exp1}), 
$2n$ samples of $X$ are partitioned into two independent datasets $\{ X_i\}_{i=1}^n$ and $\{ X_i'\}_{i=1}^n$, each having $n$ samples. The kernel MMD (squared) distance has the empirical estimator \cite{gretton2012kernel}
\[
  \gamma^2_k(P_n,P_n') =  \frac{1}{n^2} \sum_{i,j=1}^n  k(X_i, X_j) +   \frac{1}{n^2} \sum_{i,j=1}^n k(X_i',X_j') 
  - \frac{2}{n^2} \sum_{i=1,j}^n  k(X_i, X_j'), 
\]
and then, by definition,
\begin{align*}
\ex{  \gamma^2_k(P_n,P_n') }
& =   2  ( \frac{1}{n}  \ex{k(X,X)} + (1- \frac{1}{n})   \ex{k(X,X')}
  - 2  \ex{k(X,X')}  \\
& =   \frac{2}{n } (  \ex{k(X,X)} -  \ex{k(X,X')} ).
\end{align*}
Recall that 
\[
  \gamma^2_k(P,P_n) = \int \int  k(x,x') dP(x) dP(x') +   \frac{1}{n^2} \sum_{i,j=1}^n k(X_i, X_j) 
  - \frac{2}{n} \sum_{i=1}^n  \int  k(x, X_i) dP(x),
\]
and then
\begin{align*}
 \ex{ \gamma^2_k(P,P_n)}
 & =  \ex{k(X,X')}
 +     \frac{1}{n} \ex{k(X,X)} + (1-\frac{1}{n}) \ex{k(X,X')}  
  -  2 \ex{  k( X, X')} \\
 & =  
    \frac{1}{n} (  \ex{k(X,X)} -  \ex{k(X,X')} ).
\end{align*}
Thus, if we define
\[
\hat{\Delta}_\gamma^2 \coloneqq \frac{n}{2} \gamma_k^2( P_n, P_n'),
\]
the expectation of $\hat{\Delta}_\gamma^2$ equals $ \ex{k(X,X)} -  \ex{k(X,X')} = \ex{ n \gamma^2_k(P,P_n)}$.

\subsection{Experiments in Section~\ref{sec:dependent_samples_example}}
\begin{figure}[t]
\centering
   \includegraphics[width=0.32 \textwidth]{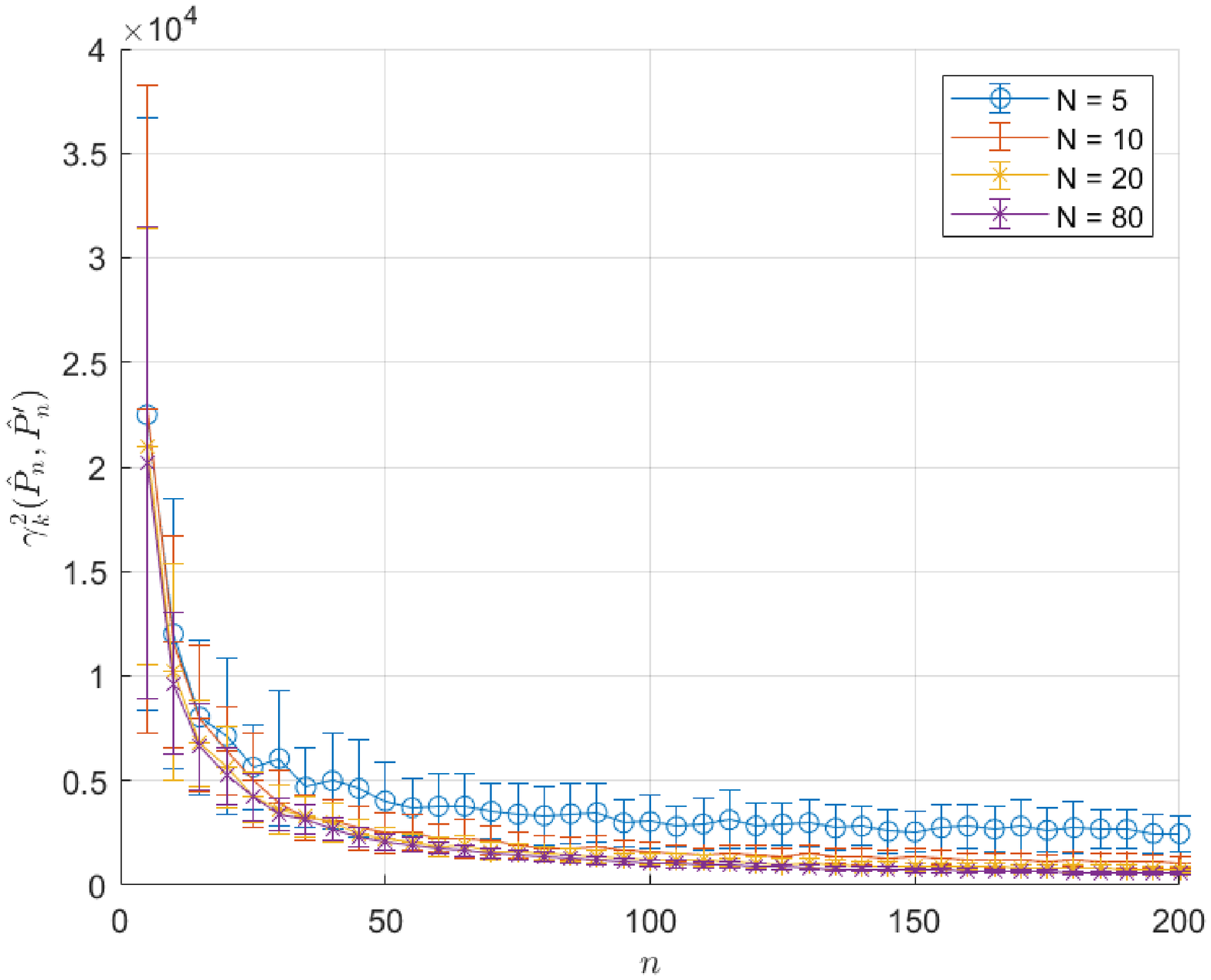}
    \includegraphics[width=0.32\textwidth]{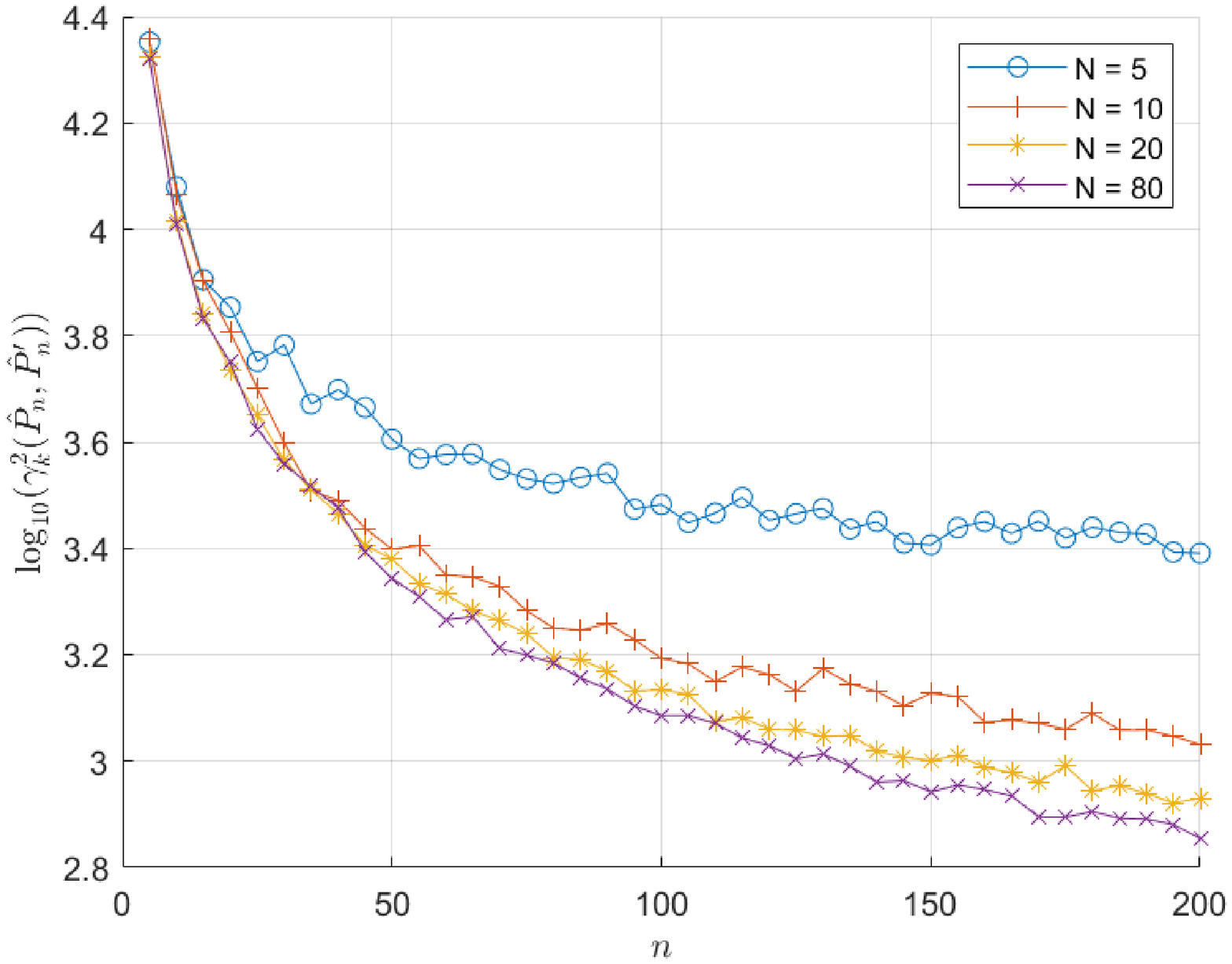}
    \includegraphics[width=0.30\textwidth]{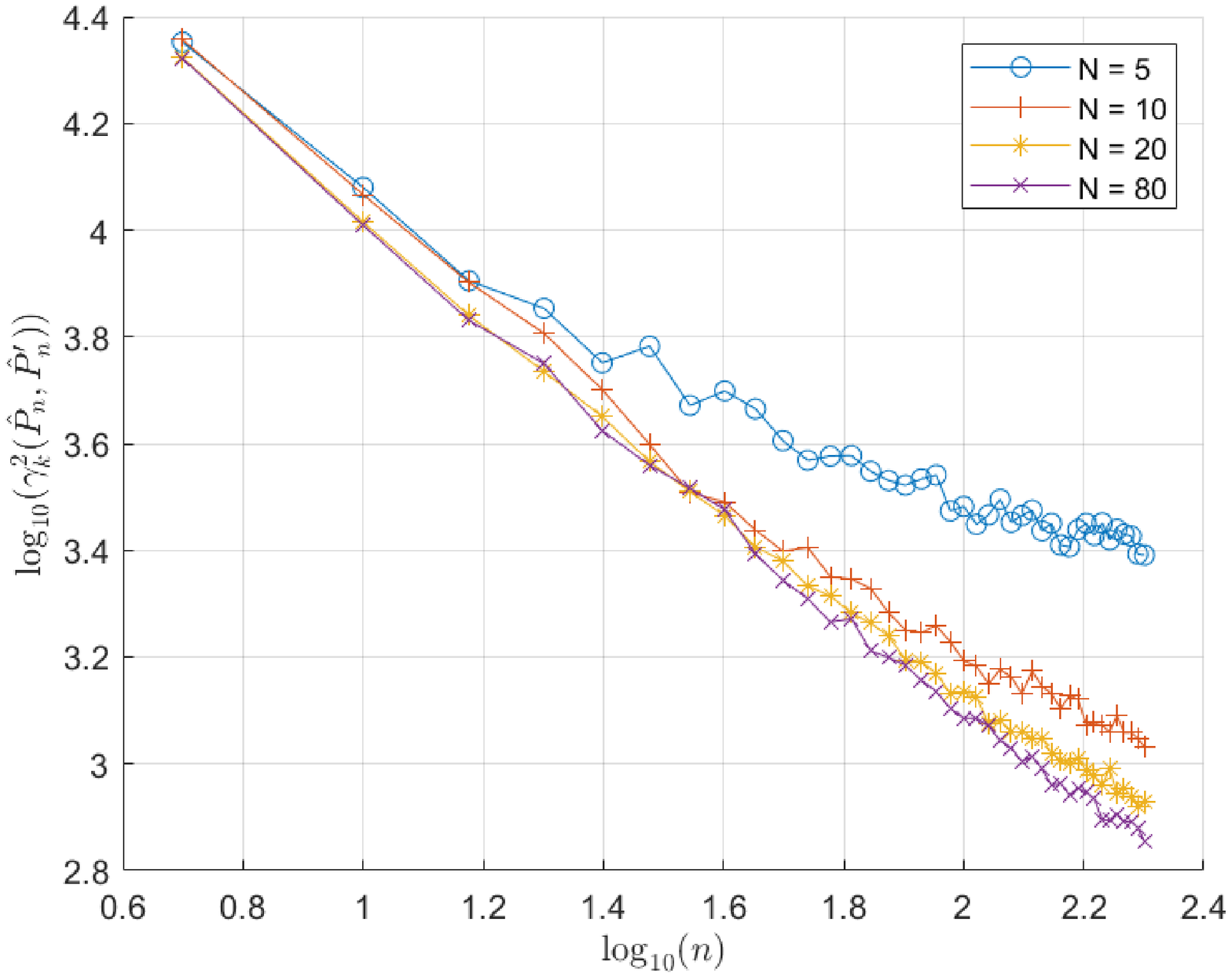}
    \caption{
    Empirical values of $\gamma_k^2 (P_n, P_n')$  as a function of $n$ for various values of $N$ for dependent samples in Example  \ref{eg:dependent-sphere}, 
    that is, the same experiment as in Figure \ref{fig:exp2}.
    (Left) Mean and standard deviation averaged over 100 realizations. (Middle) The $\log_{10}$ of the mean value in the left plot. (Right) The log-log plot of the mean value in the left plot. 
    %
%
    \label{fig:ex_1_Fig_2d_supp}}
\end{figure}

In this experiment,  $d =5$, $p=1$,  $\sigma = 1/2$ and the parameters $(\eps, \lambda)$ of the two-moment kernel are specified as in Theorem~6 with parameters $(1,0)$. 
Figure \ref{fig:exp2} in the main text plots the values of  $\gamma_{k}^{2}\left( P_n,  P^{\prime}_{n}\right)$ as a function of increasing $N$ and for various  values of $n$. 
Figure~\ref{fig:ex_1_Fig_2d_supp} plots $\gamma_{k}^{2}\left( P_n,  P^{\prime}_{n}\right)$
as a function of increasing $n$ and for various values of $N$. 
Note that in this setting, 
the typical correlation between  samples is of magnitude $1/\sqrt{N}$, 
and thus the overall dependence is not negligible when $N$ is relatively small compared to $n$.

\end{document}